\documentclass[journal]{IEEEtran}
\usepackage{amsmath,amsfonts}
\usepackage{algorithm}
\usepackage{algpseudocode}
\usepackage{array}
\usepackage[caption=false,font=normalsize,labelfont=sf,textfont=sf]{subfig}
\usepackage{textcomp}
\usepackage{stfloats}
\usepackage{float}
\usepackage{url}
\usepackage{verbatim}
\usepackage{graphicx}
\usepackage{cite}
\usepackage{threeparttable}
\usepackage{overpic}
\usepackage{rotating}
\usepackage{makecell}
\usepackage[normalem]{ulem}
\useunder{\uline}{\ul}{}
\usepackage[colorlinks,linkcolor=blue]{hyperref}
\usepackage{orcidlink}
\usepackage{tabularx}
\usepackage{booktabs}
\usepackage{colortbl} 
\usepackage{multirow}
\definecolor{low}{HTML}{D6EAF8}
\definecolor{high}{HTML}{FADBD8}
\definecolor{grey}{HTML}{EAE7E7}
\newcolumntype{C}{>{\centering\arraybackslash}X} % centered

\begin{document}

\title{DepthMaster: Taming Diffusion Models for Monocular Depth Estimation}

\author{Ziyang Song{*\thanks{*Equal contribution.}}~\orcidlink{0009-0009-6348-8713}, 
    Zerong Wang{*}~\orcidlink{0009-0001-6677-0572},
    Bo Li~\orcidlink{0000-0001-7817-0665}, 
    Hao Zhang~\orcidlink{0009-0007-1175-5918}, 
    Ruijie Zhu~\orcidlink{0000-0001-6092-0712},
    Li Liu~\orcidlink{0009-0004-3280-8490},\\
    % Jinwei Chen, \\
    Peng-Tao Jiang\textsuperscript{$^\dagger$\thanks{$^\dagger$Corresponding authors.}}~\orcidlink{0000-0002-1786-4943},
    % Tianzhu Zhang\textsuperscript{$^\dagger$ }~\orcidlink{0000-0003-0764-6106}
    Tianzhu Zhang\textsuperscript{$^\dagger$}~\orcidlink{0000-0003-0764-6106}
%     % 
% \thanks{*Equal contribution.}
\thanks{
Ziyang Song, Ruijie Zhu, Li Liu, and Tianzhu Zhang are with School of Information Science and Technology, University of Science and Technology of China (USTC), Hefei 230026, P.R.China. Contact:  
\{songziyang, ruijiezhu, liu\_li\}@mail.ustc.edu.cn;
\{tzzhang\}@ustc.edu.cn.

Zerong Wang, Bo Li, Hao Zhang, Peng-Tao Jiang are with vivo Mobile Communication Co., Ltd., Hangzhou 310030, P.R.China. 
Contact:  
\{wangzerong\}@live.com; \{libra, haozhang, pt.jiang\}@vivo.com.

This work was done during Ziyang Song's internship at vivo.
}% <-this % stops a space
% \thanks{Manuscript received August 24, 2023; revised November 1, 2023; accepted by November 14, 2023. Copyright © 20xx IEEE. Personal use of this material is permitted. However, permission to use this material for any other purposes must be obtained from the IEEE by sending an email to pubs-permissions@ieee.org.}
% \thanks{Manuscript received August 24, 2023.}
}

\markboth{IEEE TRANSACTIONS ON CIRCUITS AND SYSTEMS FOR VIDEO TECHNOLOGY,~Vol.~XX, No.~XX, XX~2025}%
{Shell \MakeLowercase{\textit{Song et al.}}: DepthMaster}

\maketitle

\begin{abstract}
Monocular depth estimation within the denoising diffusion paradigm demonstrates impressive generalization ability but suffers from low inference speed.
{Recent methods adopt a single-step deterministic paradigm to improve inference efficiency. 
However, uncritically applying the generative features from diffusion models to the perceptual depth estimation task leads to suboptimal results,
% in terms of performance, visual fidelity and quality,
due to the differences between image generation and dense prediction.}
In this work, we propose DepthMaster, a single-step diffusion model designed to tame diffusion models for enhanced performance and visual quality.
First, to mitigate overfitting to texture details introduced by generative features, we propose a Feature Alignment module, which incorporates high-quality semantic features to enhance the denoising network's representation capability.
Second, to address the lack of fine-grained details in the single-step deterministic framework, we propose a Fourier Enhancement module to adaptively balance low-frequency structure and high-frequency details.
We adopt a two-stage training strategy to fully leverage the potential of the two modules.
In the first stage, we focus on learning the global scene structure with the Feature Alignment module, while in the second stage, 
we exploit the Fourier Enhancement module to improve the visual quality.
Through these efforts, our model achieves state-of-the-art performance in terms of generalization and detail preservation, outperforming other diffusion-based methods across various datasets.
Our project page can be found at  \url{https://indu1ge.github.io/DepthMaster_page}.
\end{abstract}

\begin{IEEEkeywords}
Monocular depth estimation, Zero-shot depth estimation, Diffusion models.
\end{IEEEkeywords}

\section{Introduction}
\label{introduction}
\IEEEPARstart{M}{onocular} depth estimation (MDE) has garnered considerable attention due to its simplicity, low cost, and ease of deployment.
Unlike traditional depth-sensing techniques such as LiDAR or stereo vision, MDE only requires a single RGB image as input, making it highly appealing for a wide range of applications, including autonomous driving~\cite{schon2021mgnet, wang2019pseudo, youpseudo, kong2023robodepth}, virtual reality~\cite{luo2020consistent, noraky2019low}, and image synthesis~\cite{zhang2023adding, esser2023structure}.
This versatility also presents a significant challenge: achieving exceptional generalization to effectively handle the diversity and complexity of broad-range application scenarios. 
However, this is a non-trivial task due to variants in scene layouts, depth distributions, lighting conditions, etc.

Recent research on zero-shot monocular depth estimation has evolved into two main branches: data-driven~\cite{hu2024metric3d,yang2024depth, depth_anything_v2, ranftl2021vision, bhat2023zoedepth, eftekhar2021omnidata, ranftl2020towards, yin2020diversedepth, zhu2024scaledepth} and model-driven~\cite{ke2024repurposing, fu2025geowizard, gui2024depthfm, xu2024diffusion, he2024lotus}.
{The former relies on large-scale image-depth pairs to achieve generalization across various scenes, where data collection and training process is extremely time-consuming and resource-exhausting.
In contrast, model-driven approaches aim to adapt powerful pre-trained backbones with a small amount of annotated data and shorter training schedules, particularly in the context of Stable Diffusion models~\cite{rombach2022high, esser2024scaling}.}
% which are trained on large-scale, high-quality LAION-5B~\cite{schuhmann2022laionb}. 
% 
For example, 
Marigold~\cite{ke2024repurposing} reformulates depth estimation as a denoising diffusion process, achieving impressive performance in both generalization and detail preservation.
However, the iterative denoising process results in a low inference speed.
{GenPercept~\cite{xu2024diffusion} proposes a deterministic single-step paradigm, which directly inputs RGB images and outputs depth maps,
improving inference efficiency while maintaining comparable performance.
% 
% TODO
Despite these advancements in applying diffusion models to MDE, few works have thoroughly explored how to best adapt the generative features in diffusion models for the perceptual task.}

\begin{figure*}[t]
    \centering
    \begin{overpic}[width=\textwidth]{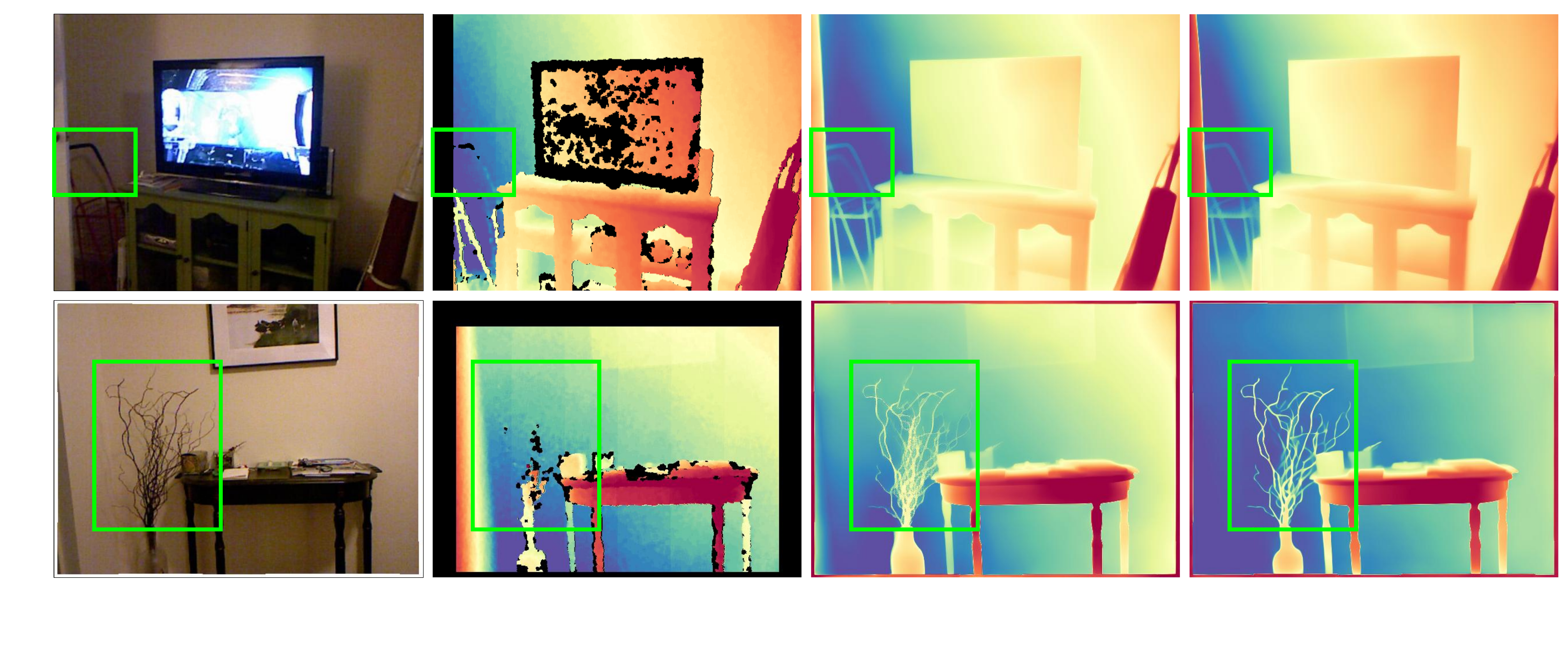}
        \put(0, 30){\begin{turn}{90}
            (a)
        \end{turn}}
        \put(0, 11.5){\begin{turn}{90}
            (b)
        \end{turn}}
        \put(13, 2){RGB}
        \put(33.5, 2){Ground Truth}
        \put(59, 2){Single-step}
        \put(85.5, 2){Ours}
        % \put(87, 0){Stage 2}
    \end{overpic}
    \vspace{-2em}
    \caption{
    % Visualization of single-step diffusion model-based methods.
    \textbf{Challenges in current single-step diffusion model-based methods.} 
    As shown in Row (a), limited by the feature representation capability of the denoising network, depth predictions of current single-step diffusion model-based methods tend to overfit texture details (shadow of the chair) and miss the real structure.
    Besides, due to the removal of the iterative process, current single-step methods suffer from blurry outputs, as shown in Row (b).
    By delicately adapting the features in diffusion models, our method achieves accurate and fine-grained depth predictions.
    }
    \label{fig:fig1}
\end{figure*}

% \subsubsection{第三段（当前任务存在的问题）}
In this work, we conduct an in-depth analysis of the feature representation within diffusion models.
% 
% First, diffusion models are typically pre-trained to reconstruct clean images from noisy inputs, emphasizing low-level texture recovery.
{First, diffusion models are typically pre-trained to reconstruct clean images from noisy inputs, which is not capable to eliminate unnecessary texture details~\cite{yu2024repa, assran2023self, lecun2022path}.}
In contrast, monocular depth estimation is a perceptual task that requires more high-level structural and semantic understanding.
Due to this fundamental mismatch, directly applying features from pre-trained diffusion models to depth estimation often leads to inaccurate and unrealistic textures in depth predictions, as illustrated in Fig.~\ref{fig:fig1}, Row (a).
Therefore, \textbf{how to enhance the model's semantic representation capacity and reduce its reliance on irrelevant details} is a key issue for taming diffusion models for depth estimation.
Furthermore, the high visual quality of diffusion models' outputs comes from the iterative refinement process.
{In the early steps, the model learns to recover the general structure, while in later steps, details are gradually refined~\cite{lee2025beta, yang2023diffusion, choi2022perception}.
When reformulated as a single-step paradigm, the model is required to capture both primary structure and fine details in a single forward pass, which often results in blurry predictions~\cite{he2024lotus, gui2024depthfm, xu2024diffusion}, as shown in Fig.~\ref{fig:fig1}, Row (b).
}
% 
% When reformulated as a single-step paradigm, the model struggles to learn both primary structure and fine details in a single forward pass, leading to blurry predictions, as shown in Fig.~\ref{fig:fig1}, Row (b).
% (see Fig.~\ref{fig:fig1}, Column~4, red boxes).
% 
Thus, \textbf{how to improve the fine-grained details in the single-step paradigm} is another crucial challenge in leveraging the generative features for depth estimation.

To address the aforementioned challenges, we propose DepthMaster, a tamed single-step diffusion model designed to enhance the generalization and detail preservation abilities of depth estimation models.
First, to enhance the representational capacity of the diffusion model, we align its features with those from a high-quality external encoder via a Feature Alignment module. 
This alignment effectively injects semantic context and alleviates overfitting to texture details.
Second, to alleviate the lack of fine-grained details due to the removal of the iterative process, we propose a Fourier Enhancement module.
It adaptively balances low-frequency structure and high-frequency details in a single forward pass, effectively simulating the learning process of the multi-step denoising.
To fully leverage the potential of the modules, we adopt a two-stage training strategy.
The first stage leverages the Feature Alignment module to learn scene structure, while the second stage incorporates the Fourier Enhancement module to improve fine-grained details.
{Through tailoring generative features and exploiting the two-stage training strategy, our method achieves impressive zero-shot performance and superior detail preservation ability, outperforming existing state-of-the-art diffusion model-based approaches, thereby narrowing the gap between data-driven and model-driven paradigms.}

The main contributions of our work are as follows:
\begin{itemize}
\item We propose \textbf{DepthMaster}, a novel approach that customizes generative features in diffusion models to suit the perceptual depth estimation task.
\item We introduce a Feature Alignment module to mitigate overfitting to texture details with high-quality external features and a Fourier Enhancement module to refine fine-grained details in the frequency domain.
\item Our method exhibits state-of-the-art zero-shot performance and superior detail preservation ability, surpassing other diffusion-based methods across various datasets.
\end{itemize}

\section{Related Work}

\subsection{Single-domain Monocular Depth Estimation}

Monocular depth estimation plays a crucial role in 3D perception.
Since Eigen et al.\cite{eigen2014depth} introduce convolutional neural networks for end-to-end training, a series of learning-based methods have been proposed, which can be broadly categorized into three branches: regression-based, classification-based, and denoising diffusion-based.
Regression-based methods directly predict continuous per-pixel depth values from a single image. 
Early approaches~\cite{laina2016deeper, hu2018squeeze} rely on deep residual networks to extract depth features.
To better leverage contextual information, Wang et al.\cite{wang2019web}, and Song et al.\cite{song2021monocular} introduce multi-scale fusion strategies.
Later, recent works~\cite{ranftl2021vision, yuan2022new, piccinelli2023idisc} incorporate Transformer to capture long-range dependencies, achieving notable improvements. 
With the advent of diffusion models, several methods~\cite{zhao2023unleashing, patni2024ecodepth, kondapaneni2024text} utilize pre-trained diffusion backbones to further enhance model performance. 
Beyond architecture innovations, recent research also explores improving model robustness in challenging scenes~\cite{tosi2024diffusion, song2023ec} and improving depth representations through auxiliary cues and multi-task learning~\cite{qi2018geonet, liu2024plane2depth, xu2018pad, chen2019towards, yin2019enforcing, shao2023nddepth}.
Classification-based approaches discretize the depth range into bins and classify each pixel accordingly.
% which discretizes the depth range into multiple bins and predicts the probability of each pixel falling into a specific bin. 
This idea is first introduced by Cao et al.\cite{cao2017estimating}. DORN\cite{fu2018deep} further treats it as an ordinal regression problem to capture the inherent ordering of depth intervals. 
To improve flexibility, AdaBins~\cite{bhat2021adabins} proposes an adaptive binning strategy that dynamically allocates depth intervals per image. BinsFormer~\cite{li2024binsformer} adopts a MaskFormer-style~\cite{Cheng_2022_CVPR} architecture for improved interaction between image features and bin representations. Building on this, LocalBins~\cite{bhat2022localbins} further refines the binning process by assigning bins at the pixel level, enabling finer-grained predictions.
Denoising diffusion-based methods iteratively denoise a noisy representation to recover the clean depth map.
DepthGen~\cite{saxena2023monocular} and DDP~\cite{ji2023ddp} first introduce this paradigm, demonstrating its effectiveness even with sparse ground truth supervision. 
DDVM~\cite{saxena2023surprising} extends the framework to broader dense prediction tasks such as optical flow. To enhance conditioning, DiffusionDepth~\cite{duan2024diffusiondepth} incorporates multi-scale features, leading to improved accuracy. Diffusion4RobustDepth~\cite{tosi2024diffusion} then leverages diffusion models to synthesize challenging images, thus enhancing model robustness. More recently, MonoDiffusion~\cite{shao2024monodiffusion} brings this paradigm into the self-supervised setting by diffusing a pre-trained teacher's predictions.
Although these single-domain methods achieve impressive results, they struggle to generalize across diverse real-world scenarios.

\subsection{Zero-shot Monocular Depth Estimation}
Estimating accurate depth maps for in-the-wild images is a crucial but challenging task due to variants in scene layouts, depth distributions, lighting conditions, etc.
Some pioneering works have tried to solve this problem, which can be primarily divided into two main branches: data-driven and model-driven.
The former focuses on collecting large-scale image-depth pairs to achieve the mapping from
RGB images to depth maps. 
To maintain training stability, these approaches opt to predict relative depth, which can already represent the scene structure.
For example, Diversedepth~\cite{yin2020diversedepth} and MiDaS~\cite{ranftl2020towards} predict affine-invariant depth to jointly train on multiple datasets and achieve good generalization across various scenarios.
On top of this, Omnidata~\cite{eftekhar2021omnidata} introduces a dataset that comprises roughly 14.5 million images and trains a robust depth estimation model following MiDaS to achieve zero-shot cross-dataset transfer.
More recently, ZoeDepth~\cite{bhat2023zoedepth} trains a relative depth estimation model and demonstrates its transferability to metric depth through fine-tuning on metric depth datasets.
We follow this strategy to predict relative depth because it is not only practical but also can be converted to metric depth easily.
DPT~\cite{ranftl2021vision} further enhances MiDaS by replacing the original CNN with a Vision Transformer.
Depth Anything~\cite{yang2024depth}, then, expands the training datasets with 62 million unlabeled images to further enlarge data coverage.
However, these methods rely on large-scale datasets, making the data collection and training process time-consuming and resource-exhausting.

% \textbf{Diffusion-based depth estimation}
The second branch of research seeks to improve model generalization by leveraging powerful image priors inherited in pre-trained models, especially in the context of Stable Diffusion models, which are trained on large-scale, high-quality datasets. 
Marigold~\cite{ke2024repurposing} first explores the potential of pre-trained latent diffusion models (LDMs) for monocular depth estimation by reformulating depth estimation as a conditional denoising diffusion process.
GeoWizard~\cite{fu2025geowizard} further improves it by incorporating normal estimation to enhance the ability to capture geometric details.
To address the problem of low inference efficiency caused by iterative denoising, DepthFM~\cite{gui2024depthfm} introduces flow matching to reduce the number of sampling steps at the cost of a slight performance degradation.
More recently, GenPercept~\cite{xu2024diffusion} offers a systematic analysis of fine-tuning protocols and proposes a single-step deterministic paradigm, where only the image latent is fed into the denoising network, and the noise latent is output for depth prediction.
Amazingly, it notably reduces the inference time with comparable performance.
Lotus~\cite{he2024lotus} also exploits a single-step denoising paradigm and proposes a detail preserver branch to improve visual quality.
Despite these advancements in applying diffusion models to MDE, no work has thoroughly explored how to best adapt the generative features in diffusion models for the perceptual task.
In this work, we bridge this gap by enhancing the denoising network's representation capability and adaptively refining features in the frequency domain, leading to further improvements in both performance and visual quality.

\begin{figure*}
    \centering
    \includegraphics[width=\textwidth]{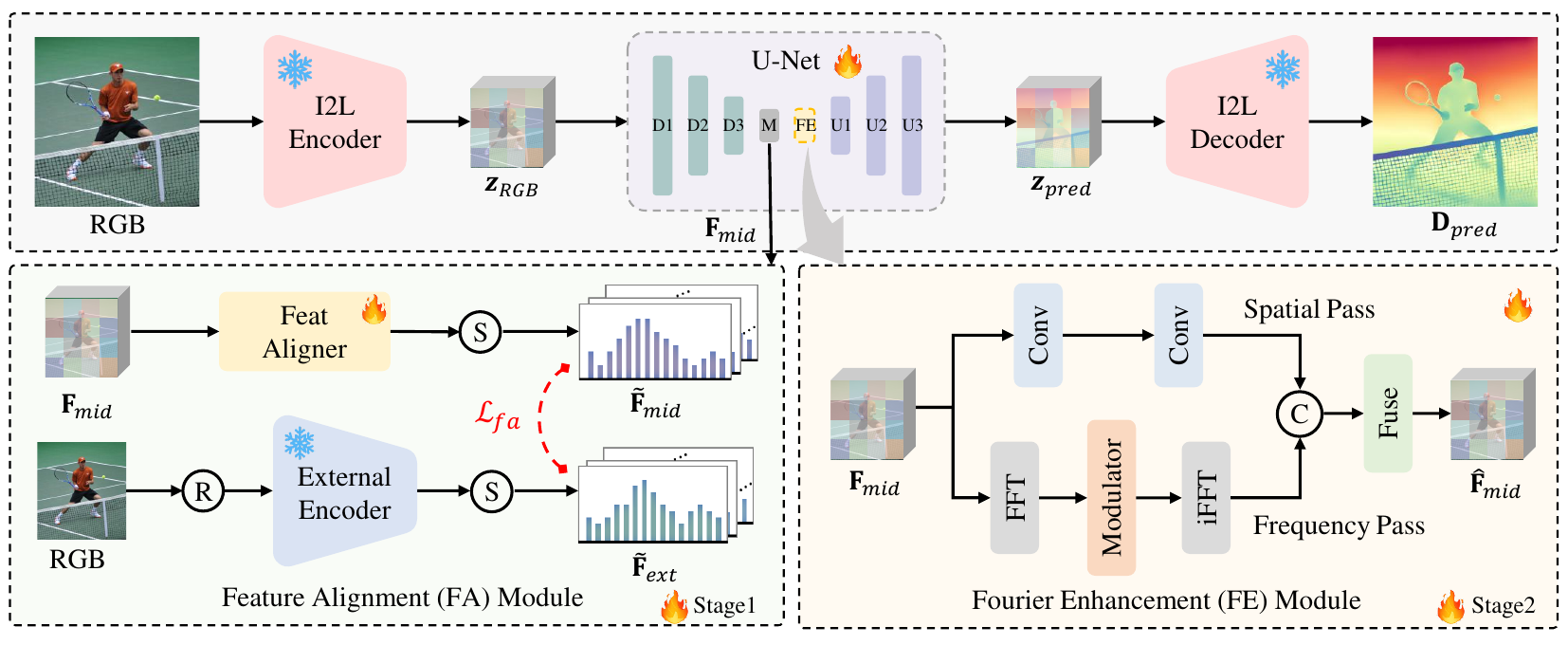}
    \caption{\textbf{The overall framework of DepthMaster.} 
    RGB is first projected into the latent space by the I2L Encoder to obtain $\textbf{z}_{RGB}$. 
    Next, the U-Net converts RGB latent to depth prediction latent $\textbf{z}_{pred}$, which is decoded back to the depth map by the I2L Decoder.
    The Feature Alignment module is applied in the first stage to align the representation of the U-Net to that of the high-quality external encoder, introducing semantic information into the diffusion model.
    In the second stage, the Fourier Enhancement module adaptively balances low-frequency structure and high-frequency details to enhance the visual quality.
    S represents the Softmax operation, R refers to the reshaping operation, and C indicates concatenation along the channel dimension.
    }
    \label{fig:framework}
\end{figure*}

\section{Method}
The overall framework is illustrated in Fig.~\ref{fig:framework}.
We begin with introducing the single-step deterministic paradigm, as detailed in Sec.~\ref{sec:rgb2disp}.
Next, we provide an in-depth analysis of the Stable Diffusion model and introduce a Feature Alignment module to enhance the representation capability of the denoising network in Sec.~\ref{sec:fa}.
To address the limitation of the single-step model's inability to capture fine-grained details, we introduce a Fourier Enhancement module in Sec.~\ref{sec:fe} and a weighted multi-directional gradient loss in Sec.~\ref{sec:grad_loss}.
Finally, we present the two-stage training strategy in Sec.~\ref{sec:2stage}.

\subsection{Deterministic Paradigm}
\label{sec:rgb2disp}
Our model is built upon Stable Diffusion v2~\cite{rombach2022high}, which is pre-trained on the large-scale LAION-5B~\cite{schuhmann2022laionb} dataset. 
The powerful image priors encoded in the model significantly assist in the depth estimation task.
To reduce computational overhead, Stable Diffusion v2 operates in the latent space by employing an image-to-latent (I2L) encoder that compresses high-resolution image data into a latent representation at $\frac{1}{8}$ the original resolution. The diffusion and denoising processes are performed within this compact latent space, and the denoised latent is subsequently decoded back into the image domain using an I2L decoder. To leverage the rich image priors learned by this model, we adopt the same paradigm and perform image perception in the latent space.
Rather than reformulating the task to fit the denoising diffusion paradigm exploited by the diffusion model, we craft the model to better adapt to the task.
Specifically, we employ a single-step deterministic transformation from image to depth, as illustrated in the upper part of Fig.~\ref{fig:framework}. 
First, the image $\textbf{I} \in \mathbb{R}^{H \times W \times 3}$ is encoded into the latent space using the I2L encoder,
% Variational Autoencoder (VAE),
denoted as $ \mathcal{E} $, and obtain the image latent 
\begin{equation}
    \textbf{z}_{RGB} = \mathcal{E}\left( \textbf{I} \right),
\end{equation}
where $\textbf{z}_{RGB} \in \mathbb{R}^{h\times w \times 4}$, $h = \frac{H}{8}$ and $w = \frac{W}{8}$. The image latent is then fed into the U-Net model, denoted as $\epsilon_\theta$, which performs scene perception and generates the corresponding depth latent. 
The timestep is set to 1, ensuring a direct conversion from image latent to depth latent:
\begin{equation}
    \textbf{z}_{pred} = \epsilon_\theta\left( \textbf{z}_{RGB}, 1 \right).
\end{equation}
Since the I2L encoder-decoder reconstructs depth maps with a negligible loss of accuracy, we only fine-tune the \mbox{U-Net}.
Instead of predicting depth, we opt to predict square-root disparity.
On the one hand, square-root disparity emphasizes the accuracy of nearby objects, which is desired by applications like autonomous driving.
On the other hand, square-root disparity leads to a more uniform distribution, thus fully releasing the capability of the input range.
The preprocessed ground truth (GT) $\textbf{D}_{GT}$ is first normalized to $\left[-1, 1\right]$ to fit the input range of the I2L encoder.
Then, we replicate it into three channels to simulate a gray-scale RGB image, which is passed through the I2L encoder to obtain the GT depth latent $\textbf{z}_{GT}$.
The training objective in the latent space is given as follows:
\begin{equation}
    \mathcal{L}_{latent} = (\textbf{z}_{GT} - \textbf{z}_{pred})^2.
\end{equation}
The final depth prediction $\textbf{D}_{pred}$ is decoded from the depth latent $\textbf{z}_{pred}$ using the I2L decoder and postprocessed by averaging channels.
The pixel-level supervision is defined as:
\begin{equation}
    \mathcal{L}_{pixel} = \Vert \textbf{D}_{GT} - \textbf{D}_{pred} \Vert ^2.
\end{equation}
By removing the iterative denoising process, the single-step deterministic paradigm significantly improves inference efficiency with comparable generalization performance.

\subsection{Feature Alignment Module}
\label{sec:fa}
Stable Diffusion v2 consists of two components: the I2L encoder-decoder and the denoising U-Net.
The I2L encoder-decoder is responsible for feature compression, which aims to reduce inference time and training costs. Trained with image reconstruction, it primarily captures low-level features. 
In contrast, the U-Net is responsible for recovering images from their noisy counterparts, enabling it to harvest scene perception and reasoning capabilities. 
% 
% However, since the U-Net is pre-trained to reconstruct clean images from noisy inputs, it tends to overemphasize fine-grained color details, 
% leading to ``pseudo-textures" rather than capturing true structure.
{However, since the U-Net is pre-trained to reconstruct clean images from noisy inputs, it is not capable to eliminate unnecessary details, 
leading to ``pseudo-textures" rather than capturing true structure.}
{Inspired by REPA~\cite{yu2024repa},
we introduce semantic regularization to enhance the U-Net's scene representation capabilities and prevent overfitting to superficial color information. }
Specifically, we incorporate a pre-trained external encoder $f$, which provides high-quality semantic feature representation.
In this work, we use DINOv2~\cite{oquab2023dinov2} as the external encoder.
For a given RGB image $\textbf{I}$, the external feature representation is $\textbf{F}_{ext} = f(\textbf{I}) \in \mathbb{R}^{h \times w \times D}$, where $h \times w$ is the spatial size and $D$ is the feature size.
Simultaneously, the U-Net encoder extracts image representation at its middle block, denoted as 
$\textbf{F}_{mid} \in \mathbb{R}^{h \times w \times C}$. 
Our goal is to align the image representation from the U-Net with the high-quality representation from the external encoder. 
Since the two representations lie in different spaces, we use a Multi-Layer Perceptron $h_\phi$ to project $\textbf{F}_{mid}$ into the feature space of $\textbf{F}_{ext}$, yielding the transformed representation $\bar{\textbf{F}}_{mid} = h_\phi(\textbf{F}_{mid})$.

To enforce feature alignment, we minimize the distance between the two feature distributions. The feature alignment loss is defined as follows:
\begin{equation}
    \mathcal{L}_{fa}(\textbf{F}_{ext}, \bar{\textbf{F}}_{mid}) = dist(\textbf{F}_{ext}, \bar{\textbf{F}}_{mid}),
\end{equation}
where $dist(\cdot, \cdot)$ measures the distance between the two feature distributions. 
In this work, we utilize the Kullback-Leibler (KL) divergence. 
Specifically, the features are first normalized along the feature dimension with the \mbox{Softmax} operation \mbox{S}, obtaining the distribution in the latent space. We then minimize the KL divergence between the two feature distributions:
\begin{equation}
    dist(\textbf{F}_{ext}, \bar{\textbf{F}}_{mid}) = kl\_div(\tilde{\textbf{F}}_{ext}, \tilde{\textbf{F}}_{mid}),
\end{equation}
where $\tilde{\textbf{F}}_{ext}$ and $\tilde{\textbf{F}}_{mid}$ represents the normalized features.
Through feature alignment, the U-Net learns more semantically meaningful representation, improving its generalization ability while avoiding overfitting to low-level details.
{Notably, the external encoder is a training-only component and removed at inference, incurring no additional test-time cost.}

\subsection{Fourier Enhancement Module}
\label{sec:fe}
The single-step paradigm effectively speeds up the inference process by avoiding multi-step iterations and multi-run integration.
{However, the fine-grained characteristic of the diffusion models' outputs typically arises from the iterative refinement process, where details are gradually refined in the later steps. 
When reformulating in the single-step paradigm, models are observed to struggle in capturing both structure and details, leading to blurry depth maps~\cite{gui2024depthfm, he2024lotus, xu2024diffusion}.
Recent studies~\cite{si2024freeu, ye2024diffusionedge, yang2023diffusion} demonstrate that frequency domain analysis provides an effective way to improve the detail quality of adapted diffusion models.
Inspired by this line of research, we propose a Fourier Enhancement module, which operates in the frequency domain to simulate the iterative refinement process's focus on different bands, as illustrated in the right-bottom of Fig.~\ref{fig:framework}.
}
% 
% Consequently, the single-step model suffers from blurry predictions.
% 
% To alleviate this problem, we propose a Fourier Enhancement module to refine high-frequency details, as illustrated in the right-bottom of Fig.~\ref{fig:framework}.
% % 
% Inspired by DiffusionEdge~\cite{ye2024diffusionedge}, the module operates in the frequency domain to simulate the iterative refinement process's focus on different bands.
% 
Specifically, the Fourier Enhancement module is composed of two components: a spatial pass for general structure capture and a frequency pass for detail enhancement.
In the frequency pass, the hidden state from the U-Net's middle block $\textbf{F}_{mid} \in \mathbb{R}^{h \times w \times C}$ is first transformed into the frequency domain using a 2D Fast Fourier Transform (FFT), yielding $\textbf{F}_{mid_f}$.
To adaptively balance information across different frequency bands, a modulator comprised of convolution and activation layers is applied to $\textbf{F}_{mid_f}$.
The enhanced feature is then transformed back to the spatial domain using an inverse 2D  Fast Fourier Transform (iFFT):
\begin{equation}
    \textbf{F}_f = {\rm iFFT}(\sigma(Conv({\rm FFT}(\textbf{F}_{mid})))),
\end{equation}
% we perform convolution in
% 
where $\sigma$ refers to the activation layer. Next, we concatenate the feature from the spatial pass $\textbf{F}_s$ with that from the frequency pass $\textbf{F}_f$ and perform a convolution operation to obtain the final enhanced feature $\hat{\textbf{F}}_{mid}$:
\begin{equation}
    \hat{\textbf{F}}_{mid} = Conv(\textbf{F}_s \Vert \textbf{F}_f),
\end{equation}
where $\Vert$ denotes the concatenation operator.
By operating in the frequency domain, our model adaptively balances low-frequency structural features and high-frequency detail features within a single forward pass, effectively improving the visual quality of depth predictions.

\subsection{Weighted Multi-directional Gradient Loss}
\label{sec:grad_loss}
To further enhance the sharpness of depth predictions, we propose a weighted multi-directional gradient loss function to capture detailed edge information on depth maps in all directions. 
Specifically, for the ground truth depth $\textbf{D}_{GT}$ and depth prediction $\textbf{D}_{pred}$, we compute gradients $\textbf{G}_{GT} \in \mathbb{R}^{H \times W \times 4}$ and $\textbf{G}_{pred} \in \mathbb{R}^{H \times W \times 4}$ in horizontal, vertical and diagonal directions. 
At edges where the foreground and background meet, gradient values are typically much larger than those within local structures. 
These dominant differences can overwhelm the gradient loss, leading to unstable training and suboptimal solutions. 
To mitigate this problem, we employ a modified Huber loss~\cite{huber1992robust}, defined as follows:
\begin{equation}
\mathcal{L}_{h} =
\left\{
\begin{array}{lr}
\delta \cdot \left| \textbf{G}_{GT} - \textbf{G}_{pred} \right|,
 &  if \left| \textbf{G}_{GT} - \textbf{G}_{pred} \right| \le \delta, \\
\frac{1}{2}\left( \textbf{G}_{GT} - \textbf{G}_{pred} \right)^2 + \frac{1}{2}\delta^2, & otherwise,
\end{array}
\right.
\end{equation}
where $\delta$ controls the threshold at which the loss transitions from quadratic to linear, reducing the influence of outliers caused by large gradient differences at the foreground-background interface.
Since the depth values are scaled to the interval \mbox{[-1, 1]}, the gradient loss corresponding to most foreground-background interface points is multiplied by a value less than 1,
thus reducing their proportion in the total gradient loss.
This adjustment allows our model to focus on the fine details not only at foreground-background interfaces but also within local structures.
% , as shown in Fig.~\ref{fig:wild_vis}.
% 
\subsection{Two-stage Training Curriculum}
\label{sec:2stage}

\begin{algorithm}[t]
\caption{The First Stage Training Process}
\label{alg:stage1}
\begin{algorithmic}[1]
\State \textbf{Input:} Training set $\mathcal{D} = \{\textbf{I}_n, \textbf{D}_n\}_{n=1}^N$, I2L encoder $\mathcal{E}$, U-Net $\epsilon_\theta$, external encoder $f$, feature aligner $h_\phi$, loss balancing factor $\lambda_{fa}$, first-stage training iteration $T_{stage1}$
\State \textbf{Initialization:} 
\State \quad Load $\mathcal{E}$, $\epsilon_\theta$ from Stable Diffusion v2
\State \quad Randomly initialize $h_\phi$
\For{ iteration t $\in$ \{1, $\cdots$, $T_{stage1}$\}}
    % \For{$(\textbf{I}, \textbf{D})$}
   \State $(\textbf{I}, \textbf{D})$ $\leftarrow$ Sampler($\mathcal{D}$)
   \State $\textbf{z}_{pred}, \textbf{F}_{mid} = \epsilon_\theta(\mathcal{E}(\textbf{I}))$
   \State $\textbf{F}_{ext} = f(\textbf{I})$
   \State $\textbf{z}_{GT} = \mathcal{E}(\textbf{D})$
   \State $\mathcal{L}_{stage1} = \mathcal{L}_{latent}(\textbf{z}_{pred}, \textbf{z}_{GT}) + \lambda_{fa}\mathcal{L}_{fa}(\textbf{F}_{ext}, \bar{\textbf{F}}_{mid})$
   \State Update $\epsilon_\theta$ and $h_\phi$ by back-propagation
\EndFor
\State \textbf{Output:} Optimized $\epsilon_\theta$
\end{algorithmic}
\end{algorithm}

\begin{algorithm}[t]
\caption{The Second Stage Training Process}
\label{alg:stage2}
\begin{algorithmic}[1]
\State \textbf{Input:} Training set $\mathcal{D} = \{\textbf{I}_n, \textbf{D}_n\}_{n=1}^N$, I2L encoder $\mathcal{E}$, I2L decoder $\mathcal{D}$, U-Net with Fourier Enhancement module $\epsilon_\psi$, loss balancing factor $\lambda_{h}$, second-stage training iteration $T_{stage2}$
\State \textbf{Initialization:} 
\State \quad Load $\mathcal{E}$, $\mathcal{D}$ from Stable Diffusion v2
\State \quad Load $\epsilon_\psi$ from the first stage model
% \State \quad Randomly initialize $h_\phi$
\For{ iteration t $\in$ \{1, $\cdots$, $T_{stage2}$\}}
    % \For{$(\textbf{I}, \textbf{D})$}
   \State $(\textbf{I}, \textbf{D})$ $\leftarrow$ Sampler($\mathcal{D}$)
   \State $\textbf{D}_{pred} = \mathcal{D}(\epsilon_\psi(\mathcal{E}(\textbf{I})))$
   \State $\textbf{G}_{GT} = \nabla(\textbf{D})$
   \State $\textbf{G}_{pred} = \nabla(\textbf{D}_{pred})$
   \State $\mathcal{L}_{stage2} = \mathcal{L}_{pixel}(\textbf{D}_{pred}, \textbf{D}) + \lambda_{h}\mathcal{L}_{h}(\textbf{G}_{GT}, \textbf{G}_{pred})$
   \State Update $\epsilon_\psi$ by back-propagation
\EndFor
\State \textbf{Output:} Optimized $\epsilon_\psi$
\end{algorithmic}
\end{algorithm}

% TODO: 两阶段的原因需要再说明一下
Since the depth reconstruction accuracy of the I2L encoder-decoder is sufficiently high, we focus on fine-tuning the U-Net.
Experiments reveal that latent-space supervision helps the model to better capture the overall scene structure, while pixel-level supervision improves fine-grained details but introduces distortions in the global structure. 
Based on these observations, we propose a two-stage training strategy. 
In the first stage, our goal is to train a model that can robustly generalize across diverse scenarios. To achieve this, we apply constraints in the latent space and incorporate the Feature Alignment module to enhance the model's scene perception capability, with the training process illustrated in Algorithm~\ref{alg:stage1}.
The training objective of the first stage is as follows:
\begin{equation}
    \mathcal{L}_{stage1} = \mathcal{L}_{latent} + \lambda_{fa} \mathcal{L}_{fa},
\end{equation}
where $\lambda_{fa}$ is set to 1.
In the second stage, we aim to optimize the model's performance on detail preservation. 
To balance structure and detail information, we incorporate the Fourier Enhancement module.
After obtaining depth predictions, we apply constraints at the pixel level and introduce the weighted multi-directional gradient loss to enhance edge sharpness, with the training process illustrated in Algorithm~\ref{alg:stage2}. 
The total objective function for the second stage is as follows:
\begin{equation}
    \mathcal{L}_{stage2} = \mathcal{L}_{pixel}+ \lambda_{h} \mathcal{L}_{h},
\end{equation}
where $\mathcal{L}_{pixel}$ is the pixel MSE loss and $\lambda_{h}$ is set to 0.001.
Benefiting from the two-stage training strategy, our model achieves both accurate structure capture and sharp edge preservation.

\section{Experiments}

\begin{table*}[!t]
\vspace{-1em}
\caption{\textbf{Quantitative comparison} with state-of-the-art zero-shot affine-invariant monocular depth estimation methods. 
The upper part lists data-driven methods and the lower part presents those based on diffusion models.
All metrics are in percentage terms with ``\textbf{bold}" best and ``\underline{underline}" second best. ``*" stands for the results reproduced by Lotus~\cite{he2024lotus}.
% Our method outperforms other diffusion model-based methods.
}
    \label{tab:overall}
\resizebox{\textwidth}{!}{
\begin{tabular}{l|@{\hspace{4pt}}c@{\hspace{5pt}}c@{\hspace{5pt}}c|@{\hspace{5pt}}c@{\hspace{3pt}}c@{\hspace{5pt}}c@{\hspace{3pt}}c@{\hspace{5pt}}c@{\hspace{3pt}}c@{\hspace{5pt}}c@{\hspace{3pt}}c@{\hspace{5pt}}c@{\hspace{3pt}}c@{\hspace{8pt}}c}
\toprule
\multirow{2}{*}{{Method}} & \multirow{2}{*}{Year} & \multirow{2}{*}{{\makecell{Training \\ Data}}} & \multirow{2}{*}{{{\makecell{FLOPs \\ (G)}}}} & \multicolumn{2}{c}{{NYUv2}} & \multicolumn{2}{c}{{KITTI}} & \multicolumn{2}{c}{{ETH3D}} & \multicolumn{2}{c}{{ScanNet}} & \multicolumn{2}{c}{{DIODE}} & \multirow{2}{*}{{\makecell{Avg.\\ Rank}}} \\
                                 &               &                          & &{AbsRel$\downarrow$}   & {$\delta_1$}$\uparrow$    & {AbsRel$\downarrow$}   & {$\delta_1$}$\uparrow$    & {AbsRel$\downarrow$}   & {$\delta_1$}$\uparrow$    & {AbsRel$\downarrow$}    & {$\delta_1$}$\uparrow$     & {AbsRel$\downarrow$}   & {$\delta_1$}$\uparrow$    &                                     \\ \midrule
\multicolumn{14}{l}{\textcolor[rgb]{0.7, 0.7, 0.7}{Data-driven methods}} \\                        
{DiverseDepth}~\cite{yin2020diversedepth}  & {arXiv'20}                   & {320K}               & {207.8}                             & {11.7}              & {87.5}     & {19.0}              & {70.4}         & {22.8}              & {69.4}           & {10.9}               & {88.2}            & {37.6}              & {63.1}           & {11}                                 \\
MiDaS~\cite{ranftl2020towards}  &  {TPAMI'20}                         & 2M                & {184.6}                            & 11.1               & 88.5      & 23.6              & 63.0          & 18.4              & 75.2           & 12.1                & 84.6            & 33.2              & 71.5           & 10.4                                 \\
LeReS~\cite{yin2021learning}   & {CVPR'21}                         & 354K                  & {190.4}                         & 9.0               & 91.6        & 14.9              & 78.4       & 17.1              & 77.7           & 9.1                & 91.7            & 27.1              &  76.6     & 7.8                                 \\
Omnidata~\cite{eftekhar2021omnidata}     & {ICCV'21}                    & 12.2M                & {195.4}                   & 7.4               & 94.5           & 14.9              & 83.5           & 16.6              & 77.8           & 7.5         & 93.6            & 33.9              & 74.2           & 7.2                                 \\
DPT~\cite{ranftl2021vision}       & {ICCV'21}                       & 1.4M                & {449.2}                     & 9.8               & 90.3       & 10.0        & 90.1        & 7.8       &  94.6     & 8.2                & 93.4            & {\ul 18.2}              & 75.8           & 6.1                                 \\
HDN~\cite{zhang2022hierarchical}   & {NIPS'22}                           & 300K                  & {195.4}                       &  6.9         & 94.8     & 11.5              & 86.7      & 12.1              & 83.3           & 8.0                & 93.9      & 24.6     &  78.0  & 5.3                           \\
{MonoDiffusion}~\cite{shao2024monodiffusion}               & {TCSVT'24}             & {39K}                  & {38.8}                 & {24.5}      & {55.0}  & {16.4}      & {73.4}  & {20.0}      & {65.1}  & {22.4}       & {61.3}   &  {34.6}        & {50.5}           & {11.4}                        \\
{Diffusion4RobustDepth}~\cite{tosi2024diffusion}               & {ECCV'24}             & {30K}                  & {159.5}                 & {8.9}      & {92.4}  & {13.0}      & {83.6}  & {17.8}      & {89.3}  & {8.8}       & {93.0}   &  {28.1}        & {71.4}           & {7.7}                        \\
{Metric3D}~\cite{yin2023metric3d}   & {ICCV'23}             & {8M}                  & {225.4}                 & {5.8}      & {96.3}  & {{\ul 5.8}}      & {{\ul 97.0}}  & {6.6}      & {96.0}  & {7.4}       & {94.1}   &  {22.4}        & {{\ul 78.5}}           & {3.3}                        \\
{Depth Anything}~\cite{yang2024depth}   & {CVPR'24}             & {62M}                  & {586.0}                 & {\textbf{4.3}}      & {\textbf{98.1}}  & {7.6}      & {{94.7}}  & {6.5}      & {{ 97.2}}  & {{\ul 4.2}}       & {98.0}   &  {27.7}        & {75.9}           & {3.3}                        \\
Depth Anything V2~\cite{depth_anything_v2}      & {NIPS'24}             & 63.5M                  & {1816.5}              & {\ul 4.4}      & {\ul 97.9}      & 7.5      & 94.8  & {\ul 6.2}      & {\ul 98.0}  & 4.3       & {\ul 98.1}   &  26.0        & 75.9           & {\ul 3.1}                        \\
{Metric3Dv2}~\cite{hu2024metric3d}      & {TPAMI'24}             & {16M}                  & {520.6}                 & {\textbf{4.3}}      & {\textbf{98.1}}  & {\textbf{4.3}}      & {\textbf{98.2}}  & {\textbf{4.2}}      & {\textbf{98.3}}  & {\textbf{2.2}}       & {\textbf{99.4}}   &  {\textbf{13.6}}        & {\textbf{89.5}}           & {\textbf{1.0}}                     \\
\midrule
\multicolumn{14}{l}{\textcolor[rgb]{0.7, 0.7, 0.7}{Model-driven methods}} \\
Marigold~\cite{ke2024repurposing}  & {CVPR'24}      & 74K                & {5196.5}                         & 5.5               & 96.4       & 9.9               & 91.6           & 6.5               & 96.0           & 6.4                & 95.1            & 30.8              & 77.3           & 4.3                                 \\
GeoWizard~\cite{fu2025geowizard}   & {ECCV'25}    & 280K                  & {5247.3}                  & {\ul 5.2}         & 96.6           & 9.7               & 92.1           & 6.4         & {\ul 96.1}     & 6.1                & 95.3            & 29.7              & \textbf{79.2}  & {\ul 2.9}                           \\
DepthFM~\cite{gui2024depthfm}    & {AAAI'25}     & 74K                & {2487.4}                  & 6.0               & 95.5            & {\ul 9.1}         & 90.2             & 6.5               & 95.4           & 6.6                & 94.9            & {\ul 22.4}        & {\ul 78.5}     & 4.5                                 \\
GenPercept~\cite{xu2024diffusion}  & {ICLR'25}    & 74K                   & {2142.8}               & 5.6               & 96.0               & 9.9               & 90.4          & {\ul 6.2}                & 95.8            & 6.2*               & 96.1*           & 35.7              & 75.6           & 4.4                                 \\
Lotus~\cite{he2024lotus}    & {ICLR'25}       & 59K              & {2142.8}                       & 5.3               & {\ul 96.7}       & 9.3               & {\ul 92.8}   & 6.8               & 95.3           & {\ul 6.0}          & {\ul 96.3}      & 22.8              & 73.8           & 3.5                                 \\
DepthMaster (Ours)     & {-}       & 74K                     & {2147.1}                 & \textbf{5.0}      & \textbf{97.2} & \textbf{8.2}      & \textbf{93.7}  & \textbf{5.3}      & \textbf{97.4}  & \textbf{5.5}       & \textbf{96.7}   & \textbf{21.5}     & 77.6           & \textbf{1.2}                        \\ \bottomrule
\end{tabular}
}

\end{table*}

\begin{figure*}[!t]
    \centering
    \begin{overpic}[width=\textwidth]{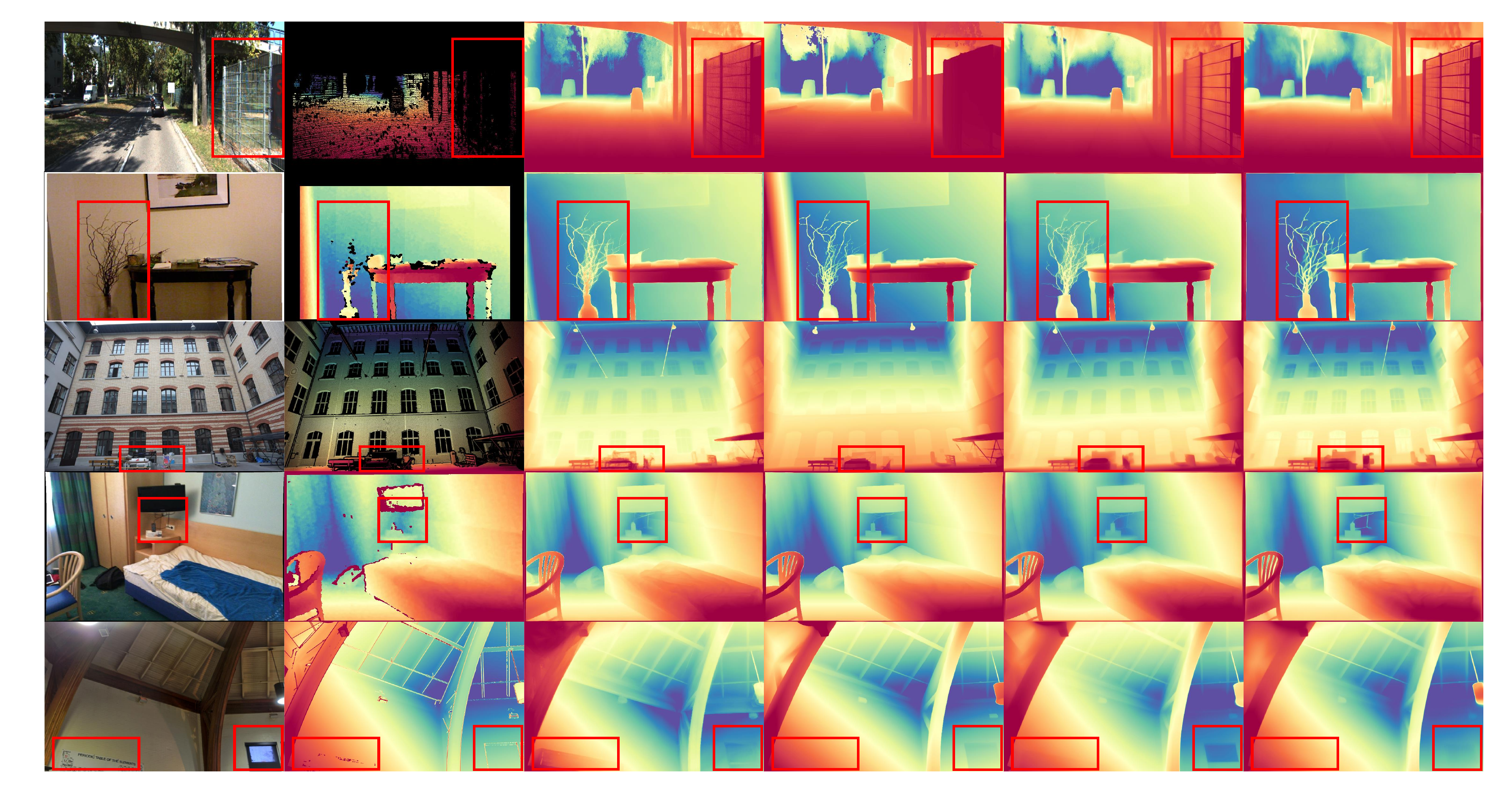}
        \put(0, 45){\begin{turn}{90}
            KITTI
        \end{turn}}
        \put(0, 34.5){\begin{turn}{90}
            NYUv2
        \end{turn}}
        \put(0, 24.8){\begin{turn}{90}
            ETH3D
        \end{turn}}
        \put(0, 14.5){\begin{turn}{90}
            ScanNet
        \end{turn}}
        \put(0, 4.7){\begin{turn}{90}
            DIODE
        \end{turn}}
        \put(9.3, 0){RGB}
        \put(21.6, 0){Ground Truth}
        \put(37.5, 0){Marigold~\cite{ke2024repurposing}}
        \put(55.8, 0){DAv2~\cite{depth_anything_v2}}
        \put(71, 0){Lotus~\cite{he2024lotus}}
        \put(89, 0){Ours}
    \end{overpic}
    \caption{\textbf{Qualitative comparison across different datasets} with zero-shot monocular depth estimation methods. Our model demonstrates excellent detail preservation and structure capture capabilities. Benefiting from the Feature Alignment module, our model avoids overfitting to textures.
    }
    \label{fig:bench_vis}
\end{figure*}

\subsection{Implementation Details}
Our model is based on Stable Diffusion v2~\cite{rombach2022high}, with text conditioning disabled.
In the first stage, we train our model for 20k iterations using the Adam~\cite{loshchilov2017decoupled} optimizer with a learning rate of $3\times10^{-5}$.
In the second stage, we reduce the learning rate to $3\times10^{-6}$ and train for an additional 10k iterations.
To achieve a batch size of 32, we exploit gradient accumulation.
Training the first stage takes approximately 30 hours, while fine-tuning the model in the second stage requires an additional 30 hours, both on a single NVIDIA H800 GPU.
Additionally, we apply random horizontal flipping augmentation to enhance the diversity of training datasets.

\subsection{Datasets}
\textbf{Training Datasets.}
We train our model on two synthetic datasets: Hypersim~\cite{roberts2021hypersim} and Virtual KITTI~\cite{cabon2020virtual}.
Hypersim is a high-fidelity dataset covering 461 indoor scenes with rich textures and geometry, generated using realistic 3D rendering techniques.
We use the depth annotations and corresponding RGB images to train our model.
Following Marigold~\cite{ke2024repurposing}, we transform the original depth values relative to the focal point into depth values relative to the focal plane.
The official split with around 54K samples is used with the training resolution of 480$\times$640.
Virtual KITTI is a synthetic outdoor dataset that serves as a variant of the original KITTI~\cite{geiger2013vision} dataset, providing a wide range of road scenes under diverse lighting, weather, and traffic conditions.
Unlike KITTI, Virtual KITTI is generated with a 3D simulator and provides dense depth annotations.
We train on approximately 20k samples with a resolution of 1216 $\times$ 352 and set the far plane to 80 meters.
The two datasets are mixed in a ratio of 9:1.

\textbf{Evaluation Datasets.}
We evaluate our model's zero-shot performance on 5 real datasets.
NYU-Depth-V2 (NYUv2)~\cite{silberman2012indoor} and ScanNet~\cite{dai2017scannet} are indoor datasets commonly used for evaluating depth estimation methods.
We use the official test split with 654 images for NYUv2 and the split proposed by Marigold with 800 images for ScanNet.
KITTI~\cite{geiger2013vision} is a street-scene dataset captured with equipment mounted on a moving vehicle. We follow the Eigen split~\cite{eigen2014depth}, which consists of 652 images.
ETH3D~\cite{schops2017multi} and DIODE~\cite{vasiljevic2019diode} are two real datasets containing both indoor and outdoor images.
For evaluation, we use the splits in Marigold to evaluate on 454 samples from ETH3D and 771 samples from DIODE.
For zero-shot evaluation, we report absolute relative error $AbsRel=\frac{1}{HW}\Sigma_{HW}\frac{\left| 
\textbf{D}_{GT} - \hat{\textbf{D}}_{pred} \right|}{\textbf{D}_{GT}}$ and accuracy metric $\delta_1=\frac{1}{HW}\Sigma_{HW}\left[ 
max\left(\frac{\hat{\textbf{D}}_{pred}}{\textbf{D}_{GT}}, \frac{\textbf{D}_{GT}}{\hat{\textbf{D}}_{pred}}\right) < 1.25 \right]$, where $\hat{\textbf{D}}_{pred}$ is the aligned depth prediction and $[\cdot]$ is Iverson bracket.
For sharp boundary evaluation, we use the F1-score proposed by DepthPro~\cite{bochkovskii2024depth}, which computes the recall and precision of edges in depth predictions and ground truth depth maps. 
This metric reflects the visual quality of depth predictions to some extent, with higher values indicating higher visual quality.
% More details will be exhibited in Supplementary materials.

\begin{figure*}[!t]
    \centering
    \hspace{-0.7em}
    \begin{overpic}[width=0.98\textwidth]{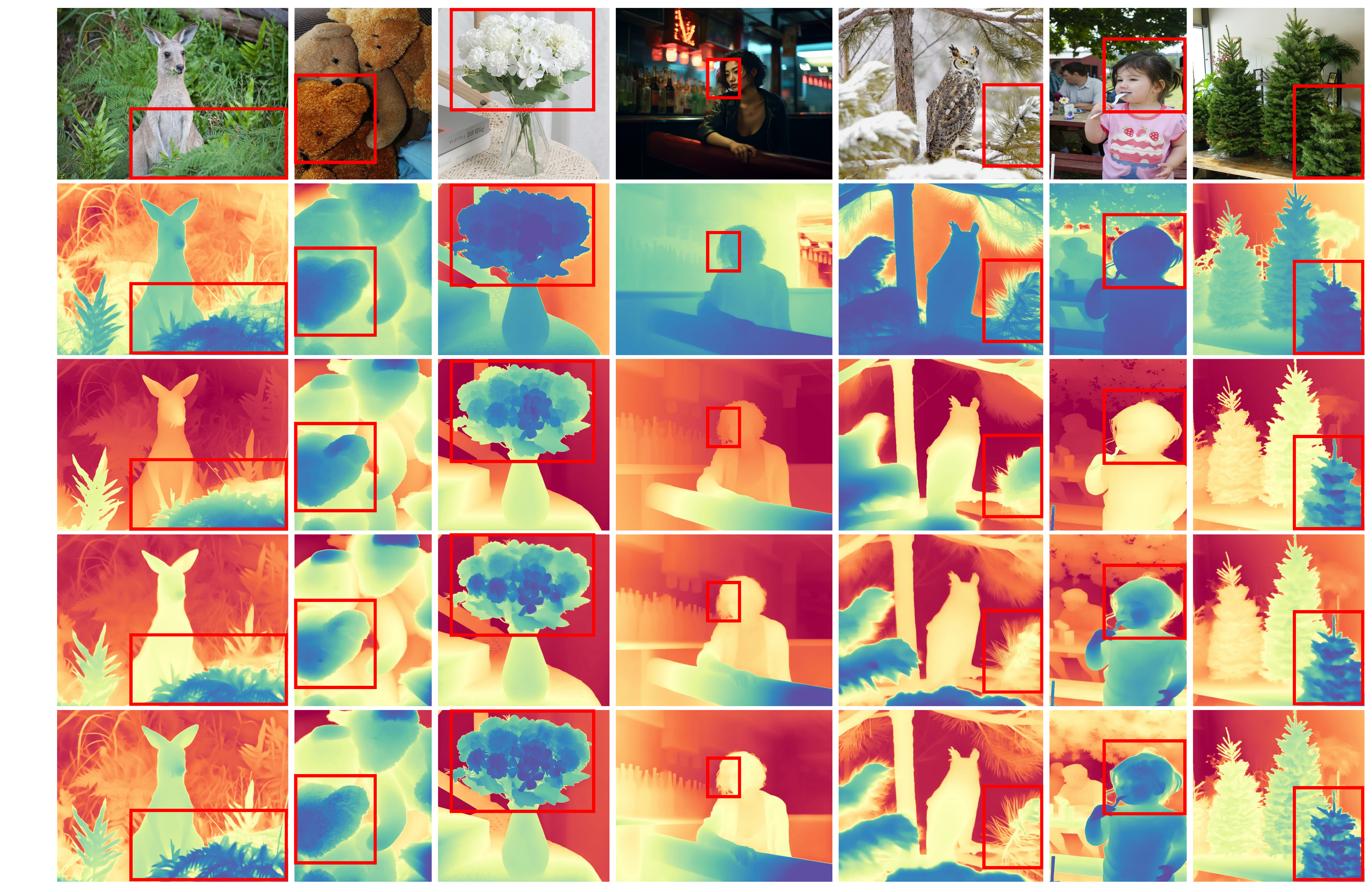}
        \put(1, 55.3){\begin{turn}{90}
            RGB
        \end{turn}}
        \put(1, 39){\begin{turn}{90}
            Marigold~\cite{ke2024repurposing}
        \end{turn}}
        \put(1, 28.5){\begin{turn}{90}
            DAv2~\cite{depth_anything_v2}
        \end{turn}}
        \put(1, 15.5){\begin{turn}{90}
            Lotus~\cite{he2024lotus}
        \end{turn}}
        \put(1, 5.2){\begin{turn}{90}
            Ours
        \end{turn}}
    \end{overpic}
    \caption{\textbf{Qualitative results on in-the-wild examples.} Our model not only recovers correct scene structure, but also exhibits fine-grained details.}
    \label{fig:wild_vis}
\end{figure*}

\subsection{Qualitative and Quantitative Comparison}

Table~\ref{tab:overall} presents a comparison of our method with other state-of-the-art (SOTA) zero-shot monocular depth estimation approaches. 
The upper part of the table lists data-driven methods, while the lower part focuses on diffusion model-based methods. 
As shown in Table~\ref{tab:overall}, despite being trained on a relatively small amount of data, diffusion model-based methods already outperform many approaches that rely on large-scale datasets.
This highlights the significant role of strong image priors encoded in diffusion models, which greatly enhance the generalization capabilities of depth estimation models. 
Our approach belongs to the diffusion model-based category.
By incorporating the specially designed Feature Alignment module and Fourier Enhancement module, 
we achieve a 17.2\% improvement over Marigold~\cite{ke2024repurposing} in AbsRel on KITTI, effectively narrowing the performance gap between diffusion model-based methods and those reliant on large-scale datasets. 
Besides, the single-step deterministic paradigm effectively improves the inference efficiency through the removal of the iterative denoising process, whose computation complexity is comparable to Depth Anything V2~\cite{depth_anything_v2}.
To better illustrate the superiority of our approach, we provide qualitative results in Fig.~\ref{fig:bench_vis}.
% 
% As highlighted in red boxes, our method excels in recovering complete structure and preserving fine-grained details, while avoiding texture overfitting commonly encountered in leveraging generative models.
{As highlighted in red boxes, the multi-step iterative diffusion model Marigold~\cite{ke2024repurposing} demonstrates fine details in thin objects, whereas the single-step Lotus~\cite{he2024lotus} often overlooks details such as thin wires, leading to relatively blurry outputs.
With the proposed Fourier Enhancement Module and the two-stage training strategy, our method effectively alleviates this issue and achieves comparable detail preservation with Marigold, while avoiding texture overfitting commonly encountered in leveraging generative models.}
Additionally, Fig.~\ref{fig:wild_vis} showcases predictions on in-the-wild examples, further demonstrating our model's remarkable generalization ability in unconstrained real-world scenarios.
These results emphasize the practical applicability and versatility of our approach, making it highly suitable for various real-world applications.
\begin{table*}[ht]
\centering
\caption{\textbf{Ablation of paradigm.} 
``I2L" means feeding depth maps into the image-to-latent (I2L) encoder-decoder and outputting reconstructed ones. 
``Denoising" refers to the multi-step denoising diffusion paradigm. 
``Iterative" indicates iterative refinement by passing through the U-Net four times in a deterministic way.
``Single-step" denotes the single-step deterministic paradigm.
``*" marks the paradigm we use.
``M.Full" is the final model.
}
    \label{tab:paradigm}
% \resizebox{0.9\textwidth}{!}{
\begin{tabular}{c|c@{\hspace{5pt}}cc@{\hspace{5pt}}cc@{\hspace{5pt}}cc@{\hspace{5pt}}cc@{\hspace{5pt}}cc@{\hspace{7pt}}c}
\toprule
\multirow{2}{*}{\textbf{Paradigm}} & \multicolumn{2}{c}{\textbf{KITTI}} & \multicolumn{2}{c}{\textbf{NYUv2}} & \multicolumn{2}{c}{\textbf{ScanNet}} & \multicolumn{2}{c}{\textbf{ETH3D}} & \multicolumn{2}{c}{\textbf{DIODE}} & \textbf{Hypersim}   & \textbf{Inference} \\
                                   & \textbf{AbsRel$\downarrow$}    & \boldmath{$\delta_1$}$\uparrow$   & \textbf{AbsRel$\downarrow$}   & \boldmath{$\delta_1$}$\uparrow$  & \textbf{AbsRel$\downarrow$}     & \boldmath{$\delta_1$}$\uparrow$    & \textbf{AbsRel$\downarrow$}    & \boldmath{$\delta_1$}$\uparrow$   & \textbf{AbsRel$\downarrow$}    & \boldmath{$\delta_1$}$\uparrow$   & \textbf{F1$\uparrow$}  &  \textbf{Time} (s)    \\ \midrule
I2L                         & -                  & -             & 1.1               & 99.5         & 0.9                 & 99.7           & -                  & -             & 8.4                & 92.4          & 0.615     & -        \\
Denoising                          & 10.4               & 90.2          & 5.7               & 96.0         & 6.9                 & 94.6           & 6.4                & 95.7          & 30.9               & 76.8          & 0.274       &  12.91    \\
Iterative                          & 10.0               & 91.1          & 5.2               & 96.7         & 5.9                 & 96.1           & 6.1                & 96.3          & 29.4               & 77.8          & 0.310   &    0.83      \\
Single-step* (M.Single)                      & 10.3               & 90.4          & 5.3               & 96.6         & 6.0                 & 96.2           & 6.5                & 95.8          & 29.9               & 77.0          & 0.304      &  0.42     \\
{M.Full}                  & {8.2}                & {93.7}          & {5.0}                & {97.2}          & {5.3}                & {97.4}          & {5.5}                 & {96.7}           & {21.5}               & {77.6}          & {0.337}  & {0.42}           \\ 
\bottomrule
\end{tabular}
% }
\end{table*}
\begin{table*}[!t]
\centering
\caption{\textbf{Ablation of depth preprocess.} Predicting disparity instead of depth results in improved performance on outdoor datasets, while using square-root disparity leads to consistent improvements across all datasets.}
    \label{tab:depth_preprocess}
% \resizebox{0.8\textwidth}{!}{
\begin{tabular}{c|c@{\hspace{5pt}}cc@{\hspace{5pt}}cc@{\hspace{5pt}}cc@{\hspace{5pt}}cc@{\hspace{5pt}}c}
\toprule
\multirow{2}{*}{\makecell{\textbf{Depth Preprocess}\\Baseline: M.Single}} & \multicolumn{2}{c}{\textbf{KITTI}} & \multicolumn{2}{c}{\textbf{NYUv2}} & \multicolumn{2}{c}{\textbf{ETH3D}} & \multicolumn{2}{c}{\textbf{ScanNet}} & \multicolumn{2}{c}{\textbf{DIODE}} \\
                                           & \textbf{AbsRel$\downarrow$}    & \boldmath{$\delta_1$}$\uparrow$   & \textbf{AbsRel$\downarrow$}   & \boldmath{$\delta_1$}$\uparrow$  & \textbf{AbsRel$\downarrow$}    & \boldmath{$\delta_1$}$\uparrow$   & \textbf{AbsRel$\downarrow$}     & \boldmath{$\delta_1$}$\uparrow$    & \textbf{AbsRel$\downarrow$}    & \boldmath{$\delta_1$}$\uparrow$   \\ 
                                           \midrule
depth (D)                                      & 10.3               & 90.4          & 5.3               & 96.6         & 6.5                & 95.8          & 6.0                 & 96.2           & 29.9               & 77.0          \\
disparity ($\frac{1}{D}$)                                       & 8.9                & 92.4          & 5.3               & 97.0         & 6.7                & 96.7          & 5.7                 & 96.3           & 22.4               & 74.0          \\
sqrt disp ($\frac{1}{\sqrt{D}}$) (M.Base)                                 & 8.7                & 93.1          & 5.1               & 97.3         & 5.5                & 97.2          & 5.8                 & 96.4           & 21.8               & 77.2          \\ 
\bottomrule
\end{tabular}
% }
\end{table*}

\subsection{Ablation Studies}
In this section, we conduct comprehensive experiments to validate the effectiveness of our designs.
We first conduct a series of additive ablations on the learning paradigm, depth preprocessing strategies, model components and the feature alignment model and its integration location.
Then we assess the role of the two-stage training strategy, the gradient loss $\mathcal{L}_{h}$, and the Fourier Enhancement module in preserving details and maintaining performance.

\textbf{Learning Paradigm.}
The ablation study of the learning paradigm is shown in Table~\ref{tab:paradigm}. 
``I2L" refers to feeding depth maps into the image-to-latent (I2L) encoder-decoder and outputting the reconstructed depth maps.\footnote{Only datasets with dense depth maps are evaluated.} 
As shown in Table~\ref{tab:paradigm}, the reconstruction accuracy of the I2L encoder-decoder is sufficiently high. 
That is to say, in the paradigm of the diffusion model, the main performance bottleneck is in the U-Net part, which is also the focus of our work.
``Denoising" refers to predicting depth maps from noise using the denoising diffusion paradigm, while ``Single-step" indicates directly predicting depth from RGB images. 
Obviously, applying the diffusion model in a deterministic manner is better suited for the discriminative task, which not only enhances the model's generalization capability but also significantly improves inference efficiency.
Actually, in the denoising diffusion paradigm, noise is progressively added to ensure outputs' diversity, which is not desirable in deterministic tasks, thus impairing the performance.
When predicting depth in an iterative deterministic way, where the U-Net's output is iteratively input to the U-Net for 4 times, the performance of the model is further improved. 
This is because the iterative paradigm aligns with the multi-step denoising process used by diffusion models, therefore, better harnessing the prior knowledge inherent in diffusion models.
However, the iterative process inevitably leads to a low inference speed, resulting in approximately twice the inference time.
{To address this issue, we adapt features in diffusion models using carefully designed modules, enabling a single-step deterministic approach that achieves zero-shot performance comparable to the iterative paradigm while maintaining nearly identical inference time with the single-step paradigm.}

\begin{figure}[t!]
    \centering
    \includegraphics[width=\linewidth]{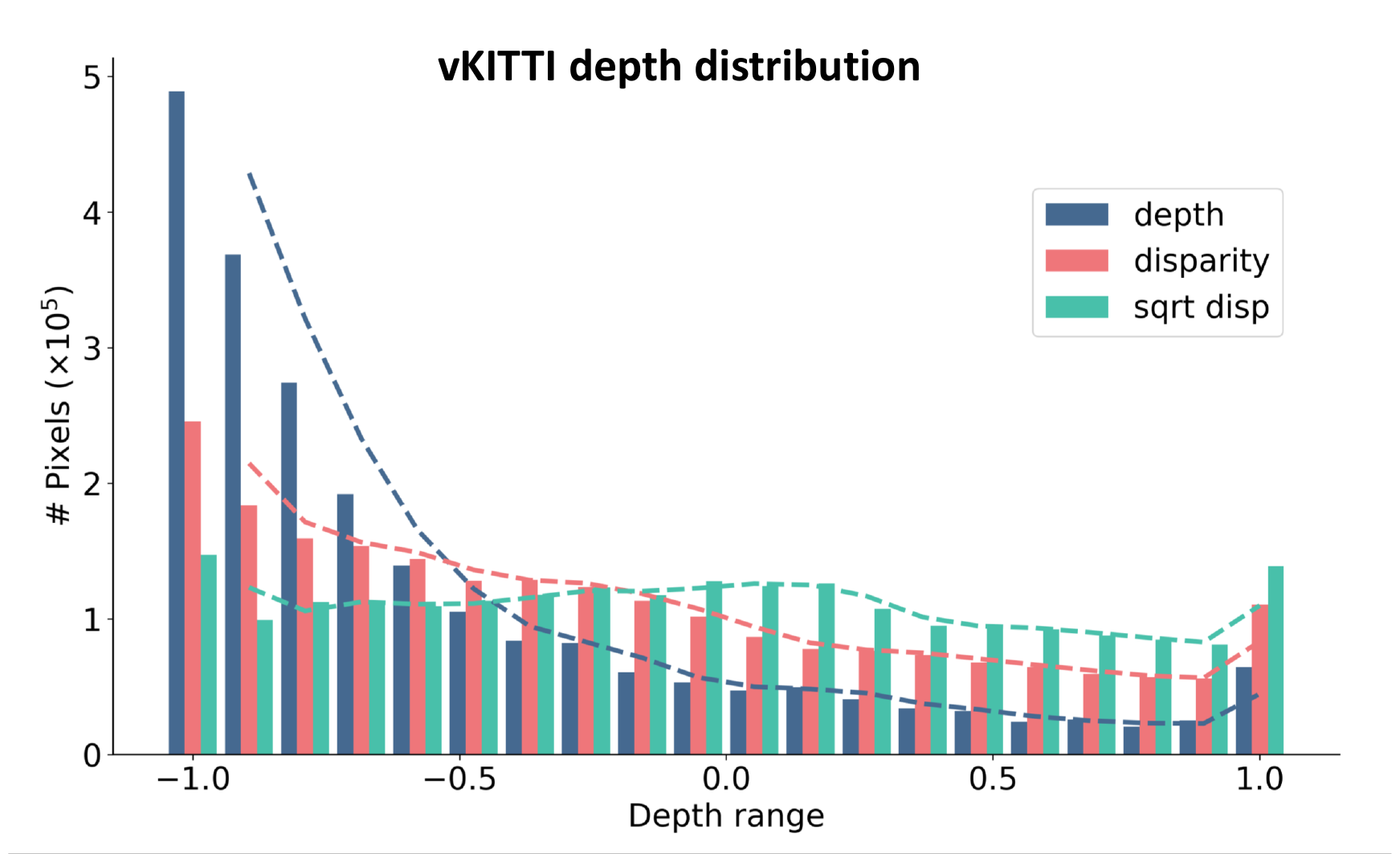}
    \caption{\textbf{Depth distribution of different depth preprocess methods} on Virtual KITTI. Square-root disparity exhibits the most uniform distribution.}
    \label{fig:distribution}
\end{figure}

\textbf{Depth Preprocess.} 
We conduct ablation studies on three different depth preprocessing methods, including depth, disparity, and square-root disparity (sqrt disp) under the single-step deterministic paradigm. 
To ensure compatibility with the input range of Stable Diffusion v2, 
the preprocessed depth maps are normalized to the range of \mbox{[-1, 1]} using percentiles. 
The results are shown in Table~\ref{tab:depth_preprocess}.
Switching from depth prediction to disparity prediction results in a notable performance improvement, particularly on outdoor and mixed indoor-outdoor datasets.
This improvement can be attributed to the fact that disparity amplifies the foreground structure, helping the model focus more on nearby objects, which is desired by outdoor applications such as autonomous driving.
Furthermore, predicting square-root disparity yields an additional performance boost, which we adopt as our baseline. 
This is because square-root disparity produces a more uniform depth distribution, as illustrated in Fig.~\ref{fig:distribution},
allowing for more efficient use of the depth range.

\begin{table*}[t!]
\centering
\caption{\textbf{Ablation studies of model components.} 
``FA" indicates the Feature Alignment module. 
``2S" denotes the two-stage training strategy.
``FE" refers to the Fourier Enhancement module.
{FA effectively improves generalization, while FE and 2S jointly enhance detail preservation,  as indicated by HyperSim F1.}
}
    \label{tab:component}
    \resizebox{\textwidth}{!}{
\begin{tabular}{c|ccc|c@{\hspace{3pt}}c@{\hspace{6pt}}c@{\hspace{3pt}}c@{\hspace{6pt}}c@{\hspace{3pt}}c@{\hspace{6pt}}c@{\hspace{3pt}}c@{\hspace{6pt}}c@{\hspace{3pt}}c@{\hspace{6pt}}c}
\toprule
\multirow{2}{*}{\makecell{\textbf{Model}\\ {Baseline: M.Base}}} & \multirow{2}{*}{\textbf{FA}} & \multirow{2}{*}{\textbf{2S}} &\multirow{2}{*}{\textbf{FE}} & \multicolumn{2}{c}{\textbf{KITTI}} & \multicolumn{2}{c}{\textbf{NYUv2}} & \multicolumn{2}{c}{\textbf{Scannet}} & \multicolumn{2}{c}{\textbf{ETH3D}} & \multicolumn{2}{c}{\textbf{DIODE}} & \textbf{HyperSim} \\
                                &                  &            &                              & \textbf{AbsRel$\downarrow$}    & \boldmath{$\delta_1$}$\uparrow$   & \textbf{AbsRel $\downarrow$}    & \boldmath{$\delta_1$}$\uparrow$   & \textbf{AbsRel$\downarrow$}     & \boldmath{$\delta_1$}$\uparrow$    & \textbf{AbsRel$\downarrow$}    & \boldmath{$\delta_1$}$\uparrow$   & \textbf{AbsRel$\downarrow$}    & \boldmath{$\delta_1$}$\uparrow$   & \textbf{F1$\uparrow$}       \\ \midrule
M.Base                        & \multicolumn{1}{l}{}         & \multicolumn{1}{l}{}         & \multicolumn{1}{l|}{}        & 8.7                & 93.1          & 5.1                & 97.3          & 5.5                 & 97.2           & 5.8                & 96.4          & 21.8               & 77.2          & 0.306             \\
M.FA                     & \checkmark        & \multicolumn{1}{l}{}                             & \multicolumn{1}{l|}{}        & 8.3                & 93.7          & 5.0                & 97.3          & 5.3                 & 97.4           & 5.5                & 96.7          & 21.6               & 77.5          & 0.309             \\ 
M.Stage2         & \checkmark           & \checkmark              & \multicolumn{1}{l|}{}                                 & 8.4                & 93.5          & 5.0                & 97.2          & 5.3                 & 97.4           & 5.5                & 96.6          & 21.6               & 77.5          & 0.330             \\
M.Full                 & \checkmark               & \checkmark                           & \checkmark                            & 8.2                & 93.7          & 5.0                & 97.2          & 5.3                 & 97.4           & 5.5                & 96.7          & 21.5               & 77.6          & 0.339             \\ \bottomrule
\end{tabular}
}
\end{table*}

\begin{figure}[t!]
    \centering
    \begin{overpic}[width=0.48\textwidth]{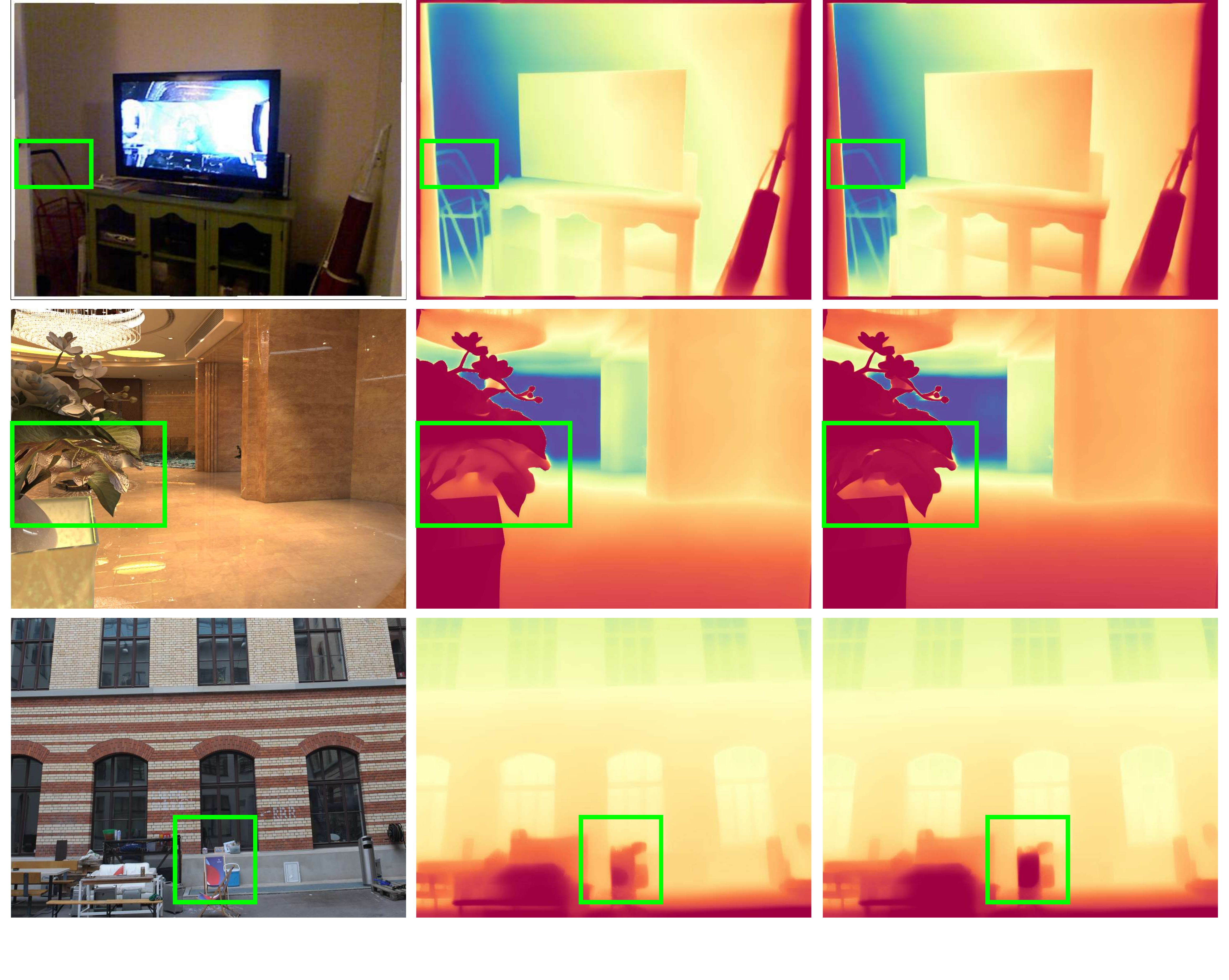}
        \put(13, 0){RGB}
        \put(45, 0){M.Base}
        \put(78, 0){M.FA}
    \end{overpic}
    \caption{{{\textbf{Overfitting to texture details of single-step diffusion model's predictions.}
    Directly applying diffusion models in a single-step discriminative way (M.Base) suffers from overfitting textures and missing the real structure.
    The Feature Alignment module (M.FA) effectively alleviates this issue.}}
        }
    \label{fig:ablation_fa}
\end{figure}

\textbf{Model Components.}
To assess the effectiveness of the proposed modules, we evaluate four model variants, with their generalization performance and edge precision summarized in Table~\ref{tab:component}.
The baseline model (M.Base) adopts a single-step deterministic paradigm, directly regressing square-root disparity from RGB images. 
Incorporating the Feature Alignment module (M.FA) yields notable improvements in generalization performance with a 4.6\% reduction in AbsRel on KITTI.
This is because the high-quality external semantic features guide the model to capture global scene structure, rather than local textures.
% 
% As illustrated in Fig~\ref{fig:ablation_fa}, 
% directly applying the generative features from diffusion models to the perceptual depth estimation task leads to overfitting to complex texture, while our Feature Alignment module effectively mitigates this issue.
{Besides, as illustrated in Fig~\ref{fig:ablation_fa}, 
directly applying generative features to the perceptual depth estimation task leads to overfitting to complex texture, since diffusion models are primarily trained for realistic image synthesis where preserving texture details is crucial. 
% In contrast, depth estimation is a perceptual task that benefits more from capturing the global scene structure. 
The proposed Feature Alignment module (M.FA) effectively eliminates the effect of unnecessary texture details by enhancing the model's semantic representation capacity.
% introducing semantic priors.
}
In the second stage, we fine-tune the first-stage model at the pixel level.
This strategy effectively improves edge precision, with an increase of 6.8\% in the F1-score, demonstrating the effectiveness of the two-stage training strategy in detail preservation.
However, due to the difficulty of learning both global structure and local details in a single forward pass, the model suffers from a slight performance drop and limited edge precision.
To address this, we introduce the Fourier Enhancement module.
Through adaptively balancing low-frequency structure and high-frequency details in the frequency domain, the Fourier Enhancement module achieves finer details and overall performance improvements.

\begin{table*}[t]
\centering
\caption{\textbf{Ablation of External Model Type in Feature Alignment module.} Introducing various external encoders can improve the generalization performance of the model, among which DINOv2 yields the greatest performance improvement.}
\label{tab:fa_encoder}
% \resizebox{\textwidth}{!}{
\begin{tabular}{c|c@{\hspace{7pt}}cc@{\hspace{7pt}}cc@{\hspace{7pt}}cc@{\hspace{7pt}}cc@{\hspace{7pt}}c}
\toprule
\multirow{2}{*}{\makecell{\textbf{External Model Type}\\ {Baseline: M.Base}}} & \multicolumn{2}{c}{\textbf{KITTI}} & \multicolumn{2}{c}{\textbf{NYUv2}} & \multicolumn{2}{c}{\textbf{ETH3D}} & \multicolumn{2}{c}{\textbf{ScanNet}} & \multicolumn{2}{c}{\textbf{DIODE}} \\
                                         & \textbf{AbsRel$\downarrow$}    & \boldmath{$\delta_1$}$\uparrow$   & \textbf{AbsRel$\downarrow$}   & \boldmath{$\delta_1$}$\uparrow$  & \textbf{AbsRel$\downarrow$}    & \boldmath{$\delta_1$}$\uparrow$   & \textbf{AbsRel$\downarrow$}     & \boldmath{$\delta_1$}$\uparrow$    & \textbf{AbsRel$\downarrow$}    & \boldmath{$\delta_1$}$\uparrow$   \\ 
                                         \midrule
M.Base                                  & 8.7                & 93.1          & 5.1               & 97.3         & 5.5                & 97.2          & 5.8                 & 96.4           & 21.8               & 77.2          \\
OpenCLIP~\cite{cherti2023reproducible}                                 & 8.5                & 93.3          & 5.0               & 97.3         & 5.4                & 97.4          & 5.6                 & 96.5           & 21.8               & 77.1          \\
AIMv2~\cite{fini2024multimodal}                                    & 8.4                & 93.4          & 5.1               & 97.3         & 5.5                & 97.3          & 5.6                 & 96.6           & 21.7               & 77.5          \\
SAM~\cite{kirillov2023segany}                                      & 8.3                & 93.5          & 5.0               & 97.3         & 5.3                & 97.5          & 5.5                 & 96.7           & 21.7               & 77.2          \\ 
{DINOv2~\cite{oquab2023dinov2}(w. REPA-E~\cite{leng2025repa})}                            & {10.4}                & {89.1}          & {6.6}               & {95.9}         & {7.2}                & {95.6}          & {7.0}                 & {95.1}           & {23.2}               & {76.0}          \\
DINOv2~\cite{oquab2023dinov2} {(M.FA)}                                  & 8.3                & 93.7          & 5.0               & 97.3         & 5.3                & 97.4          & 5.5                 & 96.7           & 21.6               & 77.5          \\
\bottomrule
\end{tabular}
% }
\end{table*}

\begin{table*}[t]
\centering
\caption{\textbf{Ablation of feature alignment location.} 
``D1", ``D2" refer to the first and second down blocks of the U-Net.
% , respectively.
``Mid" means the middle block of the U-Net.
The effectiveness of the Feature Alignment module increases as the aligned layer grows deeper.}
    \label{tab:feat_align}
\begin{tabular}{c|c@{\hspace{7pt}}cc@{\hspace{7pt}}cc@{\hspace{7pt}}cc@{\hspace{7pt}}cc@{\hspace{7pt}}c}
\toprule
\multirow{2}{*}{\makecell{\textbf{Location}\\{Baseline: M.Base}}}  & \multicolumn{2}{c}{\textbf{KITTI}} & \multicolumn{2}{c}{\textbf{NYUv2}} & \multicolumn{2}{c}{\textbf{ETH3D}} & \multicolumn{2}{c}{\textbf{ScanNet}} & \multicolumn{2}{c}{\textbf{DIODE}} \\
                                    & \textbf{AbsRel$\downarrow$}    & \boldmath{$\delta_1$}$\uparrow$   & \textbf{AbsRel$\downarrow$}   & \boldmath{$\delta_1$}$\uparrow$  & \textbf{AbsRel$\downarrow$}    & \boldmath{$\delta_1$}$\uparrow$   & \textbf{AbsRel$\downarrow$}     & \boldmath{$\delta_1$}$\uparrow$    & \textbf{AbsRel$\downarrow$}    & \boldmath{$\delta_1$}$\uparrow$   \\ 
                                   \midrule
M.Base                          & 8.7                & 93.1          & 5.1               & 97.3         & 5.5                & 97.2          & 5.8                 & 96.4           & 21.8               & 77.2          \\
D1                               & 8.5                & 93.5          & 5.0               & 97.3         & 5.3                & 97.5          & 5.6                 & 96.6           & 21.8               & 77.4          \\
D2                               & 8.4                & 93.6          & 5.1               & 97.3         & 5.4                & 97.4          & 5.5                 & 96.6           & 21.5               & 77.7          \\
Mid {(M.FA)}                             & 8.3                & 93.7          & 5.0               & 97.3         & 5.3                & 97.4          & 5.5                 & 96.7           & 21.6               & 77.5          \\ 
\bottomrule
\end{tabular}
\end{table*}

\begin{table*}[ht]
\centering
\caption{\textbf{Ablation of detail preservation.} ``pixel" indicates applying constraints at the pixel level. 
``$\mathcal{L}_h$" refers to the weighted multi-directional gradient loss.
``FE" denotes the Fourier Enhancement module. ``2S" means the two-stage training strategy. The proposed modules and training strategy effectively enhance the model's detail preservation capability. }
\label{table:fe_huber} 
% \resizebox{\textwidth}{!}{
\begin{tabular}{c|c@{\hspace{5pt}}c@{\hspace{5pt}}c@{\hspace{5pt}}c|c@{\hspace{5pt}}cc@{\hspace{5pt}}cc@{\hspace{5pt}}cc@{\hspace{5pt}}cc@{\hspace{5pt}}cc}
\toprule
\multirow{2}{*}{\makecell{\textbf{Model}\\ {Baseline: M.Base}}} & \multirow{2}{*}{\textbf{pixel}} & \multirow{2}{*}{\textbf{\boldmath{$\mathcal{L}_h$}}} & \multirow{2}{*}{\textbf{FE}} & \multirow{2}{*}{\textbf{2S}} & \multicolumn{2}{c}{\textbf{KITTI}} & \multicolumn{2}{c}{\textbf{NYUv2}} & \multicolumn{2}{c}{\textbf{ETH3D}} & \multicolumn{2}{c}{\textbf{Scannet}} & \multicolumn{2}{c}{\textbf{DIODE}} & \textbf{HyperSim} \\
                                &                                 &                                &                              &                                  & \textbf{AbsRel$\downarrow$}    & \boldmath{$\delta_1$}$\uparrow$   & \textbf{AbsRel$\downarrow$}    & \boldmath{$\delta_1$}$\uparrow$   & \textbf{AbsRel$\downarrow$}    & \boldmath{$\delta_1$}$\uparrow$   & \textbf{AbsRel$\downarrow$}     & \boldmath{$\delta_1$}$\uparrow$    & \textbf{AbsRel$\downarrow$}    & \boldmath{$\delta_1$}$\uparrow$   & \textbf{F1$\uparrow$}       \\ \midrule
M.Base                          & \multicolumn{1}{l}{}            & \multicolumn{1}{l}{}           & \multicolumn{1}{l}{}         & \textbf{}                        & 8.7                & 93.1          & 5.1                & 97.3          & 5.5                & 97.2          & 5.8                 & 96.4           & 21.8               & 77.2          & 0.306             \\
M.Pixel                         & \checkmark                               &                                &                              &                                  & 8.7                & 93.0          & 5.2                & 97.2          & 5.5                & 97.1          & 5.8                 & 96.5           & 21.8               & 77.1          & 0.307             \\
M.Huber                         & \checkmark                               & \checkmark                              &                              &                                  & 8.5                & 93.0          & 5.0                & 97.2          & 5.5                & 97.1          & 5.5                 & 96.9           & 21.6               & 77.4          & 0.308             \\
M.FE\_Huber                     & \checkmark                               & \checkmark                              & \checkmark                            &                                  & 8.3                & 93.5          & 5.1                & 97.2          & 5.3                & 97.2          & 5.5                 & 96.7           & 21.6               & 77.4          & 0.314             \\
M.Full                          & \checkmark                               & \checkmark                              & \checkmark                            & \checkmark                                & 8.2                & 93.7          & 5.0                & 97.2          & 5.3                & 97.4          & 5.5                 & 96.7           & 21.5               & 77.6          & 0.337             \\ 
\bottomrule
\end{tabular}
% }
\end{table*}

\begin{figure*}[!h]
\vspace{-1.5em}
\begin{minipage}[t]{0.49\linewidth}
\vspace{0.15em}
\begin{figure}[H]
    \begin{overpic}[width=\textwidth]{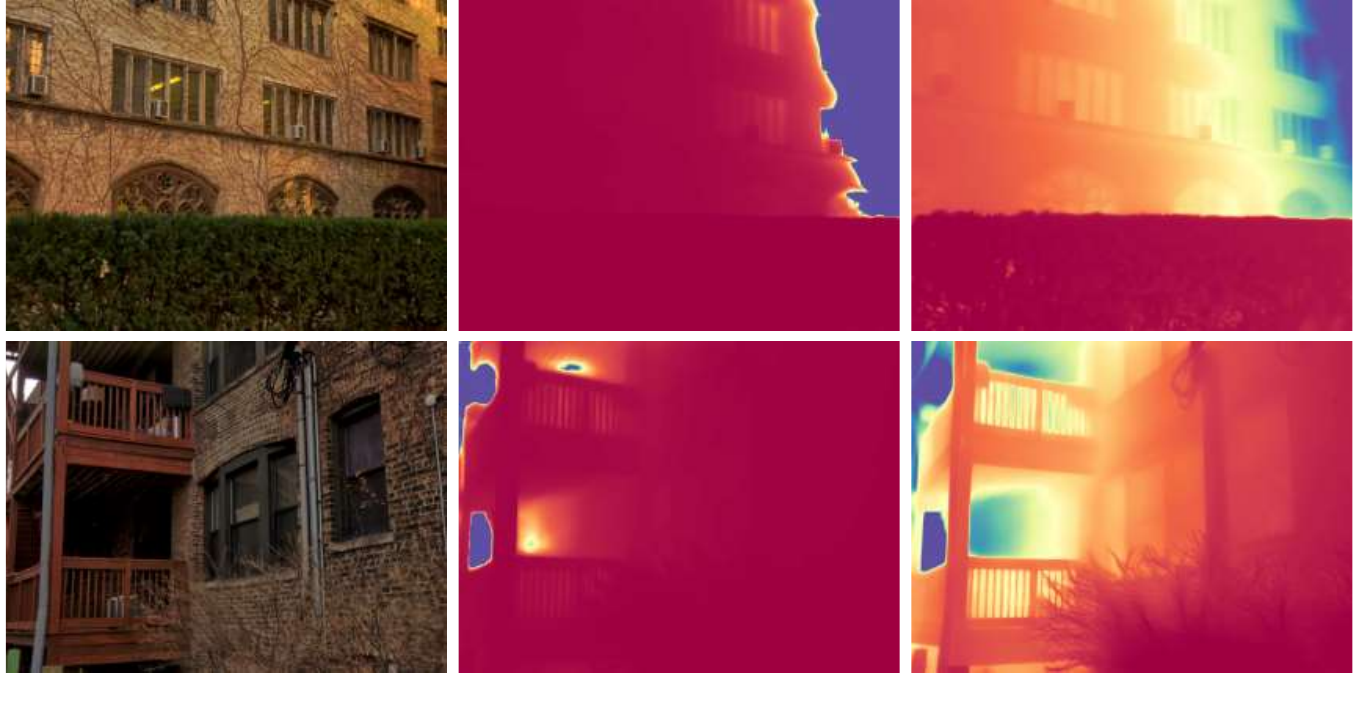}
        \put(13, -3.5){RGB}
        \put(44, -3.5){M.Pixel}
        \put(76, -3.5){M.Stage2}
    \end{overpic}
    \vspace{-0.35em}
    \vspace{-0.2em}
    \caption{{\textbf{Effect of pixel-level supervision at different stages.} Applying pixel-level constraints in the first stage (M.Pixel) introduces structural distortion, while the two-stage strategy (M.Stage2) alleviates this issue.}
    }
    \label{fig:pixel_s2}
\end{figure}
\end{minipage}
\hspace{0.8em}
\begin{minipage}[t]{0.49\linewidth}
\begin{figure}[H]
    \begin{overpic}[width=\textwidth]{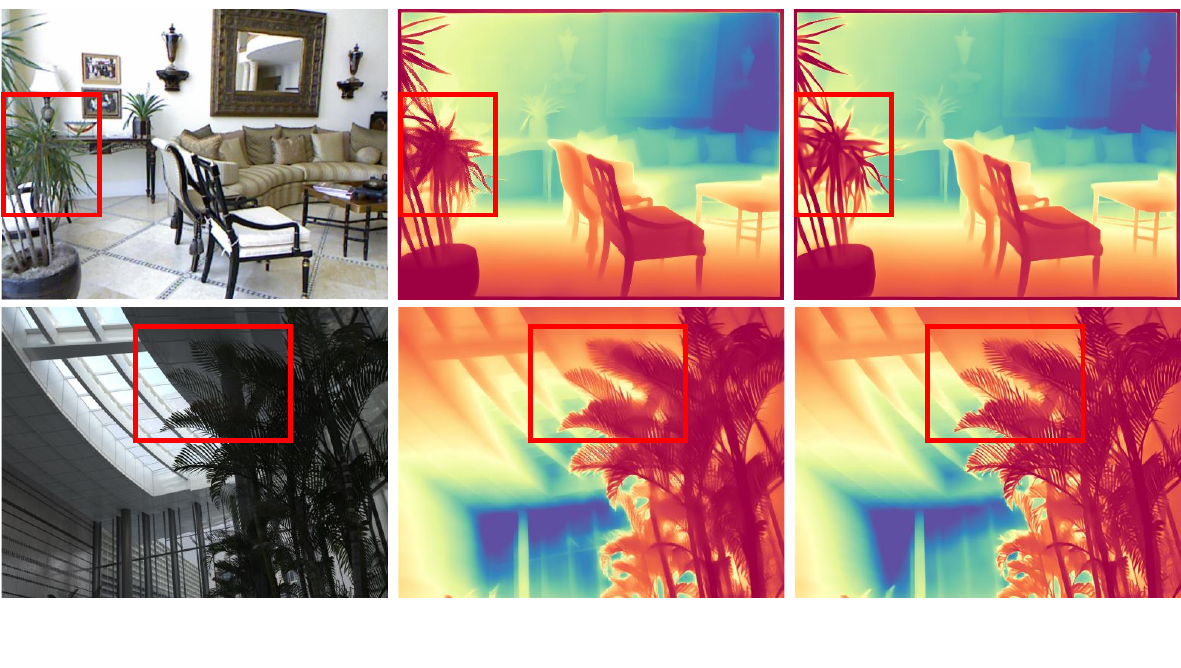}
        \put(13, 0){RGB}
        \put(45, 0){M.FA}
        \put(78, 0){M.Full}
    \end{overpic}
    \vspace{-1.35em}
    \caption{{\textbf{Visualization of predictions from two stages.} With the Fourier Enhancement Module and the two-stage training strategy, the final model (M.Full) exhibits excellent detail preservation ability.}}
    \label{fig:ablation_s2}
    \vspace{-1em}
\end{figure}
\end{minipage}
\end{figure*}

\textbf{Feature Alignment module.}
We conduct ablation studies on different external encoders and feature alignment locations in the Feature Alignment module.
As shown in Table~\ref{tab:fa_encoder}, the high-quality features of these external encoders can effectively modulate the features of the diffusion model and bring gains in zero-shot performance. 
Among them, DINOv2 has consistent performance improvements on various datasets.
{We also provide comparison with end-to-end training with REPA-E~\cite{leng2025repa}, which leads to performance degradation. This may be caused by insufficient training data and the disruption of the VAE's depth encoding capability when fine-tuned only on RGB features.}
The results of the ablation study on feature alignment locations are shown in Table~\ref{tab:feat_align}.
The U-Net is an encoder-decoder structure with three downsampling blocks, one middle block, and three upsampling blocks. 
The prior knowledge and scene perception abilities are primarily stored in the encoder part. 
Therefore, we perform feature alignment between the DINOv2 feature and the features from the first and second downsampling blocks and the middle block, respectively.
As shown in Table~\ref{tab:feat_align}, 
the effectiveness of the Feature Alignment module increases with the depth of the layer where it is applied.
This occurs because the shallow U-Net layers capture more local information and are rich in details, while the middle layer features have a better global perception, which matches the global nature of the DINOv2 features. 
When constraints are imposed on shallow layers, the detailed information will be compromised.

\textbf{Detail preservation.}
To validate the detail-preserving capability of the components we propose, we conduct a series of ablation experiments, as shown in Table~\ref{table:fe_huber}. 
``M.Base" represents our baseline. 
Directly applying constraints in the pixel space (M.pixel) and using the weighted multi-directional gradient loss (M.Huber) cannot improve the model's detail preservation capability, as indicated by the F1 metric. 
{Moreover, direct pixel-level supervision can lead to structural distortions, as shown in Fig~\ref{fig:pixel_s2}.}
This is because, in the single-step paradigm, the model is required to learn both low-frequency structural information and high-frequency details in a single forward pass, which introduces confusion during the learning process. 
When the Fourier Enhancement module is applied to adaptively enhance features in the frequency domain, both the model's generalization and fine-grained details are improved.
This demonstrates that the Fourier Enhancement module effectively mimics the iterative process's focus on different frequency bands.
Additionally, with the proposed two-stage training strategy, the model's fine-grained detail is significantly enhanced.
The second stage only need to optimize details based on the first stage model, thus simplifying the problem of capturing both structure and deatils.
Fig.~\ref{fig:ablation_s2} presents the qualitative results of the two stages, where the fine-tuned model demonstrates a remarkable detail preservation ability, highlighting the effectiveness of our strategy in improving the visual quality.

\section{\textcolor{black}{Limitations}} 
Although our method achieves performance comparable to data-driven approaches and detail preservation ability comparable to diffusion-based methods, the model's large parameter size limits its deployment on mobile devices. 
Through experimentation, we identify some redundant parameters in the U-Net, and removing these layers does not significantly affect performance. 
Therefore, reducing the model's computational cost through effective pruning and distillation techniques will be a key focus of our future work.

\section{Conclusion}
In this work, we propose DepthMaster, a method that crafts diffusion models for depth estimation. 
By incorporating the Feature Alignment module, we effectively mitigate the overfitting to texture details.
Additionally, the Fourier Enhancement module enhances fine-grained detail preservation ability bi operating in the frequency domain.
Benefiting from the careful design, DepthMaster achieves a significant boost in zero-shot performance and inference efficiency.
Extensive experiments validate the effectiveness of our approach, which achieves state-of-the-art performance in terms of generalization and detail preservation, outperforming other diffusion-based methods across various datasets.

\bibliographystyle{IEEEtran}
\bibliography{reference}

@inproceedings{wang2019pseudo,
  title={Pseudo-lidar from visual depth estimation: Bridging the gap in 3d object detection for autonomous driving},
  author={Wang, Yan and Chao, Wei-Lun and Garg, Divyansh and Hariharan, Bharath and Campbell, Mark and Weinberger, Kilian Q},
  booktitle={Proceedings of the IEEE/CVF Conference on Computer Vision and Pattern Recognition},
  pages={8445--8453},
  year={2019}
}

@inproceedings{youpseudo,
  title={Pseudo-LiDAR++: Accurate Depth for 3D Object Detection in Autonomous Driving},
  author={You, Yurong and Wang, Yan and Chao, Wei-Lun and Garg, Divyansh and Pleiss, Geoff and Hariharan, Bharath and Campbell, Mark and Weinberger, Kilian Q},
  booktitle={International Conference on Learning Representations},
  year={2019}
}

@inproceedings{schon2021mgnet,
    title={Mgnet: Monocular geometric scene understanding for autonomous driving},
    author={Sch{\"o}n, Markus and Buchholz, Michael and Dietmayer, Klaus},
    booktitle={{Proceedings of the IEEE/CVF International Conference on Computer Vision}},
    pages={15804--15815},
    year={2021}
}

@article{luo2020consistent,
  title={Consistent video depth estimation},
  author={Luo, Xuan and Huang, Jia-Bin and Szeliski, Richard and Matzen, Kevin and Kopf, Johannes},
  journal={ACM Transactions on Graphics},
  volume={39},
  number={4},
  pages={71--1},
  year={2020},
  publisher={ACM New York, NY, USA}
}

@article{noraky2019low,
  title={Low power depth estimation of rigid objects for time-of-flight imaging},
  author={Noraky, James and Sze, Vivienne},
  journal={IEEE Transactions on Circuits and Systems for Video Technology},
  volume={30},
  number={6},
  pages={1524--1534},
  year={2019},
  publisher={IEEE}
}

@inproceedings{ranftl2021vision,
  title={Vision transformers for dense prediction},
  author={Ranftl, Ren{\'e} and Bochkovskiy, Alexey and Koltun, Vladlen},
  booktitle={{Proceedings of the IEEE/CVF International Conference on Computer Vision}},
  pages={12179--12188},
  year={2021}
}

@article{li2024binsformer,
  title={BinsFormer: Revisiting Adaptive Bins for Monocular Depth Estimation},
  author={Li, Zhenyu and Wang, Xuyang and Liu, Xianming and Jiang, Junjun},
  journal={IEEE Transactions on Image Processing},
  volume={33},
  pages={3964--3976},
  year={2024},
  publisher={IEEE}
}

@inproceedings{silberman2012indoor,
  author    = {Nathan Silberman, Derek Hoiem, Pushmeet Kohli and Rob Fergus},
  title     = {Indoor Segmentation and Support Inference from RGBD Images},
  booktitle = {{Proceedings of the European Conference on Computer Vision}},
  pages={746--760},
  year = {2012}
}

@inproceedings{bhat2021adabins,
  title={Adabins: Depth estimation using adaptive bins},
  author={Bhat, Shariq Farooq and Alhashim, Ibraheem and Wonka, Peter},
  booktitle={{Proceedings of the IEEE/CVF Conference on Computer Vision and Pattern Recognition}},
  pages={4009--4018},
  year={2021}
}

@inproceedings{fu2018deep,
  title={Deep ordinal regression network for monocular depth estimation},
  author={Fu, Huan and Gong, Mingming and Wang, Chaohui and Batmanghelich, Kayhan and Tao, Dacheng},
  booktitle={{Proceedings of the IEEE/CVF Conference on Computer Vision and Pattern Recognition}},
  pages={2002--2011},
  year={2018}
}

@inproceedings{loshchilov2017decoupled,
  title={Decoupled Weight Decay Regularization},
  author={Loshchilov, Ilya and Hutter, Frank},
  booktitle={International Conference on Learning Representations},
  year={2017}
}

@article{eigen2014depth,
  title={Depth map prediction from a single image using a multi-scale deep network},
  author={Eigen, David and Puhrsch, Christian and Fergus, Rob},
  journal={{Advances in Neural Information Processing Systems}},
  volume={2},
  pages={2366--2374},
  year={2014}
}

@inproceedings{hu2018squeeze,
  title={Squeeze-and-excitation networks},
  author={Hu, Jie and Shen, Li and Sun, Gang},
  booktitle={{Proceedings of the IEEE/CVF Conference on Computer Vision and Pattern Recognition}},
  pages={7132--7141},
  year={2018}
}

@inproceedings{yin2021learning,
  title={Learning to recover 3d scene shape from a single image},
  author={Yin, Wei and Zhang, Jianming and Wang, Oliver and Niklaus, Simon and Mai, Long and Chen, Simon and Shen, Chunhua},
  booktitle={{Proceedings of the IEEE/CVF Conference on Computer Vision and Pattern Recognition}},
  pages={204--213},
  year={2021}
}

@inproceedings{yuan2022new,
  title={Neural window fully-connected crfs for monocular depth estimation},
  author={Yuan, Weihao and Gu, Xiaodong and Dai, Zuozhuo and Zhu, Siyu and Tan, Ping},
  booktitle={Proceedings of the IEEE/CVF Conference on Computer Vision and Pattern Recognition},
  pages={3916--3925},
  year={2022}
}

@inproceedings{laina2016deeper,
  title={Deeper depth prediction with fully convolutional residual networks},
  author={Laina, Iro and Rupprecht, Christian and Belagiannis, Vasileios and Tombari, Federico and Navab, Nassir},
  booktitle={International Conference on 3D vision},
  pages={239--248},
  year={2016}
}

@article{cao2017estimating,
  title={Estimating depth from monocular images as classification using deep fully convolutional residual networks},
  author={Cao, Yuanzhouhan and Wu, Zifeng and Shen, Chunhua},
  journal={IEEE Transactions on Circuits and Systems for Video Technology},
  volume={28},
  number={11},
  pages={3174--3182},
  year={2017},
  publisher={IEEE}
}

@inproceedings{qi2018geonet,
  title={Geonet: Geometric neural network for joint depth and surface normal estimation},
  author={Qi, Xiaojuan and Liao, Renjie and Liu, Zhengzhe and Urtasun, Raquel and Jia, Jiaya},
  booktitle={{Proceedings of the IEEE/CVF Conference on Computer Vision and Pattern Recognition}},
  pages={283--291},
  year={2018}
}

@inproceedings{xu2018pad,
  title={Pad-net: Multi-tasks guided prediction-and-distillation network for simultaneous depth estimation and scene parsing},
  author={Xu, Dan and Ouyang, Wanli and Wang, Xiaogang and Sebe, Nicu},
  booktitle={{Proceedings of the IEEE/CVF Conference on Computer Vision and Pattern Recognition}},
  pages={675--684},
  year={2018}
}

@inproceedings{yin2019enforcing,
  title={Enforcing geometric constraints of virtual normal for depth prediction},
  author={Yin, Wei and Liu, Yifan and Shen, Chunhua and Yan, Youliang},
  booktitle={{Proceedings of the IEEE/CVF International Conference on Computer Vision}},
  pages={5684--5693},
  year={2019}
}

@article{yin2020diversedepth,
  title={Diversedepth: Affine-invariant depth prediction using diverse data},
  author={Yin, Wei and Wang, Xinlong and Shen, Chunhua and Liu, Yifan and Tian, Zhi and Xu, Songcen and Sun, Changming and Renyin, Dou},
  journal={arXiv preprint arXiv:2002.00569},
  year={2020}
}

@article{kong2023robodepth,
  title={Robodepth: Robust out-of-distribution depth estimation under corruptions},
  author={Kong, Lingdong and Xie, Shaoyuan and Hu, Hanjiang and Ng, Lai Xing and Cottereau, Benoit and Ooi, Wei Tsang},
  journal={Advances in Neural Information Processing Systems},
  volume={36},
  pages={21298--21342},
  year={2023}
}

@article{bhat2023zoedepth,
  title={Zoedepth: Zero-shot transfer by combining relative and metric depth},
  author={Bhat, Shariq Farooq and Birkl, Reiner and Wofk, Diana and Wonka, Peter and M{\"u}ller, Matthias},
  journal={arXiv preprint arXiv:2302.12288},
  year={2023}
}

@inproceedings{eftekhar2021omnidata,
  title={Omnidata: A Scalable Pipeline for Making Multi-Task Mid-Level Vision Datasets From 3D Scans},
  author={Eftekhar, Ainaz and Sax, Alexander and Malik, Jitendra and Zamir, Amir},
  booktitle={Proceedings of the IEEE/CVF International Conference on Computer Vision},
  pages={10786--10796},
  year={2021}
}

@article{song2021monocular,
  title={Monocular depth estimation using laplacian pyramid-based depth residuals},
  author={Song, Minsoo and Lim, Seokjae and Kim, Wonjun},
  journal={IEEE Transactions on Circuits and Systems for Video Technology},
  volume={31},
  number={11},
  pages={4381--4393},
  year={2021},
  publisher={IEEE}
}

@inproceedings{wang2019web,
  title={Web stereo video supervision for depth prediction from dynamic scenes},
  author={Wang, Chaoyang and Lucey, Simon and Perazzi, Federico and Wang, Oliver},
  booktitle={International Conference on 3D vision},
  pages={348--357},
  year={2019}
}

@inproceedings{patni2024ecodepth,
  title={ECoDepth: Effective Conditioning of Diffusion Models for Monocular Depth Estimation},
  author={Patni, Suraj and Agarwal, Aradhye and Arora, Chetan},
  booktitle={Proceedings of the IEEE/CVF Conference on Computer Vision and Pattern Recognition},
  pages={28285--28295},
  year={2024}
}

@inproceedings{zhao2023unleashing,
  title={Unleashing text-to-image diffusion models for visual perception},
  author={Zhao, Wenliang and Rao, Yongming and Liu, Zuyan and Liu, Benlin and Zhou, Jie and Lu, Jiwen},
  booktitle={Proceedings of the IEEE/CVF International Conference on Computer Vision},
  pages={5729--5739},
  year={2023}
}

@inproceedings{chen2019towards,
  title={Towards scene understanding: Unsupervised monocular depth estimation with semantic-aware representation},
  author={Chen, Po-Yi and Liu, Alexander H and Liu, Yen-Cheng and Wang, Yu-Chiang Frank},
  booktitle={Proceedings of the IEEE/CVF Conference on Computer Vision and Pattern Recognition},
  pages={2624--2632},
  year={2019}
}

@article{liu2024plane2depth,
  title={Plane2Depth: Hierarchical Adaptive Plane Guidance for Monocular Depth Estimation},
  author={Liu, Li and Zhu, Ruijie and Deng, Jiacheng and Song, Ziyang and Yang, Wenfei and Zhang, Tianzhu},
  journal={IEEE Transactions on Circuits and Systems for Video Technology},
  volume={35},
  number={2},
  pages={1136--1149},
  year={2024},
  publisher={IEEE}
}

@article{ranftl2020towards,
  title={Towards robust monocular depth estimation: Mixing datasets for zero-shot cross-dataset transfer},
  author={Ranftl, Ren{\'e} and Lasinger, Katrin and Hafner, David and Schindler, Konrad and Koltun, Vladlen},
  journal={IEEE Transactions on Pattern Analysis and Machine Intelligence},
  volume={44},
  number={3},
  pages={1623--1637},
  year={2020},
  publisher={IEEE}
}

@inproceedings{rombach2022high,
  title={High-resolution image synthesis with latent diffusion models},
  author={Rombach, Robin and Blattmann, Andreas and Lorenz, Dominik and Esser, Patrick and Ommer, Bj{\"o}rn},
  booktitle={Proceedings of the IEEE/CVF Conference on Computer Vision and Pattern Recognition},
  pages={10684--10695},
  year={2022}
}

@inproceedings{he2024lotus,
    title={Lotus: Diffusion-based Visual Foundation Model for High-quality Dense Prediction},
    author={He, Jing and Li, Haodong and Yin, Wei and Liang, Yixun and Li, Leheng and Zhou, Kaiqiang and Liu, Hongbo and Liu, Bingbing and Chen, Ying-Cong},
    year={2025},
    booktitle={International Conference on Learning Representations},
}

@inproceedings{xu2024diffusion,
  title={What Matters When Repurposing Diffusion Models for General Dense Perception Tasks?},
  author={Xu, Guangkai and Ge, Yongtao and Liu, Mingyu and Fan, Chengxiang and Xie, Kangyang and Zhao, Zhiyue and Chen, Hao and Shen, Chunhua},
  year={2025},
  booktitle={International Conference on Learning Representations},
}

@inproceedings{zhang2023adding,
  title={Adding conditional control to text-to-image diffusion models},
  author={Zhang, Lvmin and Rao, Anyi and Agrawala, Maneesh},
  booktitle={Proceedings of the IEEE/CVF International Conference on Computer Vision},
  pages={3836--3847},
  year={2023}
}

@inproceedings{esser2023structure,
  title={Structure and content-guided video synthesis with diffusion models},
  author={Esser, Patrick and Chiu, Johnathan and Atighehchian, Parmida and Granskog, Jonathan and Germanidis, Anastasis},
  booktitle={Proceedings of the IEEE/CVF International Conference on Computer Vision},
  pages={7346--7356},
  year={2023}
}

@inproceedings{yang2024depth,
  title={Depth anything: Unleashing the power of large-scale unlabeled data},
  author={Yang, Lihe and Kang, Bingyi and Huang, Zilong and Xu, Xiaogang and Feng, Jiashi and Zhao, Hengshuang},
  booktitle={Proceedings of the IEEE/CVF Conference on Computer Vision and Pattern Recognition},
  pages={10371--10381},
  year={2024}
}

@article{depth_anything_v2,
  title={Depth anything v2},
  author={Yang, Lihe and Kang, Bingyi and Huang, Zilong and Zhao, Zhen and Xu, Xiaogang and Feng, Jiashi and Zhao, Hengshuang},
  journal={Advances in Neural Information Processing Systems},
  volume={37},
  pages={21875--21911},
  year={2024}
}

@inproceedings{ke2024repurposing,
  title={Repurposing diffusion-based image generators for monocular depth estimation},
  author={Ke, Bingxin and Obukhov, Anton and Huang, Shengyu and Metzger, Nando and Daudt, Rodrigo Caye and Schindler, Konrad},
  booktitle={Proceedings of the IEEE/CVF Conference on Computer Vision and Pattern Recognition},
  pages={9492--9502},
  year={2024}
}

@inproceedings{fu2025geowizard,
  title={Geowizard: Unleashing the diffusion priors for 3d geometry estimation from a single image},
  author={Fu, Xiao and Yin, Wei and Hu, Mu and Wang, Kaixuan and Ma, Yuexin and Tan, Ping and Shen, Shaojie and Lin, Dahua and Long, Xiaoxiao},
  booktitle={European Conference on Computer Vision},
  pages={241--258},
  year={2025}
}

@inproceedings{gui2024depthfm,
  title={DepthFM: Fast Generative Monocular Depth Estimation with Flow Matching},
  author={Gui, Ming and Schusterbauer, Johannes and Prestel, Ulrich and Ma, Pingchuan and Kotovenko, Dmytro and Grebenkova, Olga and Baumann, Stefan Andreas and Hu, Vincent Tao and Ommer, Bj{\"o}rn},
  booktitle={Proceedings of the AAAI Conference on Artificial Intelligence},
  volume={39},
  number={3},
  pages={3203--3211},
  year={2025}
}

@inproceedings{yu2024repa,
    title={Representation Alignment for Generation: Training Diffusion Transformers Is Easier Than You Think},
    author={Sihyun Yu and Sangkyung Kwak and Huiwon Jang and Jongheon Jeong and Jonathan Huang and Jinwoo Shin and Saining Xie},
    year={2025},
    booktitle={International Conference on Learning Representations},
}

@article{oquab2023dinov2,
  title={Dinov2: Learning robust visual features without supervision},
  author={Oquab, Maxime and Darcet, Timoth{\'e}e and Moutakanni, Th{\'e}o and Vo, Huy and Szafraniec, Marc and Khalidov, Vasil and Fernandez, Pierre and Haziza, Daniel and Massa, Francisco and El-Nouby, Alaaeldin and others},
  journal={arXiv preprint arXiv:2304.07193},
  year={2023}
}

@inproceedings{esser2024scaling,
  title={Scaling rectified flow transformers for high-resolution image synthesis},
  author={Esser, Patrick and Kulal, Sumith and Blattmann, Andreas and Entezari, Rahim and M{\"u}ller, Jonas and Saini, Harry and Levi, Yam and Lorenz, Dominik and Sauer, Axel and Boesel, Frederic and others},
  booktitle={International Conference on Machine Learning},
  year={2024}
}

@article{lecun2022path,
  title={A path towards autonomous machine intelligence version 0.9. 2, 2022-06-27},
  author={LeCun, Yann},
  journal={Open Review},
  volume={62},
  number={1},
  pages={1--62},
  year={2022}
}

@inproceedings{assran2023self,
  title={Self-supervised learning from images with a joint-embedding predictive architecture},
  author={Assran, Mahmoud and Duval, Quentin and Misra, Ishan and Bojanowski, Piotr and Vincent, Pascal and Rabbat, Michael and LeCun, Yann and Ballas, Nicolas},
  booktitle={Proceedings of the IEEE/CVF Conference on Computer Vision and Pattern Recognition},
  pages={15619--15629},
  year={2023}
}

@article{fini2024multimodal,
    title={Multimodal Autoregressive Pre-training of Large Vision Encoders},
    author={Enrico Fini and Mustafa Shukor and Xiujun Li and Philipp Dufter and Michal Klein and David Haldimann and Sai Aitharaju and Victor Guilherme Turrisi da Costa and Louis Béthune and Zhe Gan and Alexander T Toshev and Marcin Eichner and Moin Nabi and Yinfei Yang and Joshua M. Susskind and Alaaeldin El-Nouby},
    journal={arXiv preprint arXiv:2411.14402},
    year={2024}
}

@inproceedings{cherti2023reproducible,
  title={Reproducible scaling laws for contrastive language-image learning},
  author={Cherti, Mehdi and Beaumont, Romain and Wightman, Ross and Wortsman, Mitchell and Ilharco, Gabriel and Gordon, Cade and Schuhmann, Christoph and Schmidt, Ludwig and Jitsev, Jenia},
  booktitle={Proceedings of the IEEE/CVF Conference on Computer Vision and Pattern Recognition},
  pages={2818--2829},
  year={2023}
}

@article{schuhmann2022laionb,
  title={Laion-5b: An open large-scale dataset for training next generation image-text models},
  author={Schuhmann, Christoph and Beaumont, Romain and Vencu, Richard and Gordon, Cade and Wightman, Ross and Cherti, Mehdi and Coombes, Theo and Katta, Aarush and Mullis, Clayton and Wortsman, Mitchell and others},
  journal={Advances in Neural Information Processing Systems},
  volume={35},
  pages={25278--25294},
  year={2022}
}

@article{zhang2022hierarchical,
  title={Hierarchical normalization for robust monocular depth estimation},
  author={Zhang, Chi and Yin, Wei and Wang, Billzb and Yu, Gang and Fu, Bin and Shen, Chunhua},
  journal={Advances in Neural Information Processing Systems},
  volume={35},
  pages={14128--14139},
  year={2022}
}

@inproceedings{bochkovskii2024depth,
  author     = {Aleksei Bochkovskii and Ama\"{e}l Delaunoy and Hugo Germain and Marcel Santos and
               Yichao Zhou and Stephan R. Richter and Vladlen Koltun},
  title      = {Depth Pro: Sharp Monocular Metric Depth in Less Than a Second},
  booktitle  = {International Conference on Learning Representations},
  year       = {2025}
}

@inproceedings{roberts2021hypersim,
  title={Hypersim: A photorealistic synthetic dataset for holistic indoor scene understanding},
  author={Roberts, Mike and Ramapuram, Jason and Ranjan, Anurag and Kumar, Atulit and Bautista, Miguel Angel and Paczan, Nathan and Webb, Russ and Susskind, Joshua M},
  booktitle={Proceedings of the IEEE/CVF International Conference on Computer Vision},
  pages={10912--10922},
  year={2021}
}

@article{cabon2020virtual,
  title={Virtual kitti 2},
  author={Cabon, Yohann and Murray, Naila and Humenberger, Martin},
  journal={arXiv preprint arXiv:2001.10773},
  year={2020}
}

@article{geiger2013vision,
  title={Vision meets robotics: The kitti dataset},
  author={Geiger, Andreas and Lenz, Philip and Stiller, Christoph and Urtasun, Raquel},
  journal={The International Journal of Robotics Research},
  volume={32},
  number={11},
  pages={1231--1237},
  year={2013}
}

@inproceedings{dai2017scannet,
  title={Scannet: Richly-annotated 3d reconstructions of indoor scenes},
  author={Dai, Angela and Chang, Angel X and Savva, Manolis and Halber, Maciej and Funkhouser, Thomas and Nie{\ss}ner, Matthias},
  booktitle={Proceedings of the IEEE/CVF Conference on Computer Vision and Pattern Recognition},
  pages={5828--5839},
  year={2017}
}

@inproceedings{schops2017multi,
  title={A multi-view stereo benchmark with high-resolution images and multi-camera videos},
  author={Schops, Thomas and Schonberger, Johannes L and Galliani, Silvano and Sattler, Torsten and Schindler, Konrad and Pollefeys, Marc and Geiger, Andreas},
  booktitle={Proceedings of the IEEE/CVF Conference on Computer Vision and Pattern Recognition},
  pages={3260--3269},
  year={2017}
}

@article{vasiljevic2019diode,
  title={Diode: A dense indoor and outdoor depth dataset},
  author={Vasiljevic, Igor and Kolkin, Nick and Zhang, Shanyi and Luo, Ruotian and Wang, Haochen and Dai, Falcon Z and Daniele, Andrea F and Mostajabi, Mohammadreza and Basart, Steven and Walter, Matthew R and others},
  journal={arXiv preprint arXiv:1908.00463},
  year={2019}
}

@inproceedings{kirillov2023segany,
  title={Segment anything},
  author={Kirillov, Alexander and Mintun, Eric and Ravi, Nikhila and Mao, Hanzi and Rolland, Chloe and Gustafson, Laura and Xiao, Tete and Whitehead, Spencer and Berg, Alexander C and Lo, Wan-Yen and others},
  booktitle={Proceedings of the IEEE/CVF International Conference on Computer Vision},
  pages={4015--4026},
  year={2023}
}

@inproceedings{ye2024diffusionedge,
  title={DiffusionEdge: Diffusion Probabilistic Model for Crisp Edge Detection},
  author={Ye, Yunfan and Xu, Kai and Huang, Yuhang and Yi, Renjiao and Cai, Zhiping},
  booktitle={Proceedings of the AAAI Conference on Artificial Intelligence},
  volume={38},
  number={7},
  pages={6675--6683},
  year={2024}
}

@article{zhu2024scaledepth,
  title={Scaledepth: Decomposing metric depth estimation into scale prediction and relative depth estimation},
  author={Zhu, Ruijie and Wang, Chuxin and Song, Ziyang and Liu, Li and Zhang, Tianzhu and Zhang, Yongdong},
  journal={arXiv preprint arXiv:2407.08187},
  year={2024}
}

@incollection{huber1992robust,
  title={Robust estimation of a location parameter},
  author={Huber, Peter J},
  booktitle={Breakthroughs in statistics: Methodology and distribution},
  pages={492--518},
  year={1992}
}

@ARTICLE{shao2024monodiffusion,
  author={Shao, Shuwei and Pei, Zhongcai and Chen, Weihai and Sun, Dingchi and Chen, Peter CY and Li, Zhengguo},
  journal={IEEE Transactions on Circuits and Systems for Video Technology}, 
  title={MonoDiffusion: self-supervised monocular depth estimation using diffusion model},
  year={2024},
  volume={35},
  number={4},
  pages={3664-3678},
  doi={10.1109/TCSVT.2024.3509619}}

@inproceedings{bhat2022localbins,
  title={Localbins: Improving depth estimation by learning local distributions},
  author={Bhat, Shariq Farooq and Alhashim, Ibraheem and Wonka, Peter},
  booktitle={European Conference on Computer Vision},
  pages={480--496},
  year={2022}
}

@InProceedings{Cheng_2022_CVPR,
    author    = {Cheng, Bowen and Misra, Ishan and Schwing, Alexander G. and Kirillov, Alexander and Girdhar, Rohit},
    title     = {Masked-Attention Mask Transformer for Universal Image Segmentation},
    booktitle = {Proceedings of the IEEE/CVF Conference on Computer Vision and Pattern Recognition},
    month     = {June},
    year      = {2022},
    pages     = {1290-1299}
}

@article{saxena2023monocular,
  title={Monocular depth estimation using diffusion models},
  author={Saxena, Saurabh and Kar, Abhishek and Norouzi, Mohammad and Fleet, David J},
  journal={arXiv preprint arXiv:2302.14816},
  year={2023}
}

@inproceedings{piccinelli2023idisc,
  title={idisc: Internal discretization for monocular depth estimation},
  author={Piccinelli, Luigi and Sakaridis, Christos and Yu, Fisher},
  booktitle={Proceedings of the IEEE/CVF Conference on Computer Vision and Pattern Recognition},
  pages={21477--21487},
  year={2023}
}

@inproceedings{shao2023nddepth,
  title={Nddepth: Normal-distance assisted monocular depth estimation},
  author={Shao, Shuwei and Pei, Zhongcai and Chen, Weihai and Wu, Xingming and Li, Zhengguo},
  booktitle={Proceedings of the IEEE/CVF International Conference on Computer Vision},
  pages={7931--7940},
  year={2023}
}

@inproceedings{kondapaneni2024text,
  title={Text-image alignment for diffusion-based perception},
  author={Kondapaneni, Neehar and Marks, Markus and Knott, Manuel and Guimar{\~a}es, Rog{\'e}rio and Perona, Pietro},
  booktitle={Proceedings of the IEEE/CVF Conference on Computer Vision and Pattern Recognition},
  pages={13883--13893},
  year={2024}
}

@article{song2023ec,
  title={EC-Depth: Exploring the consistency of self-supervised monocular depth estimation in challenging scenes},
  author={Song, Ziyang and Zhu, Ruijie and Wang, Chuxin and Deng, Jiacheng and He, Jianfeng and Zhang, Tianzhu},
  journal={arXiv preprint arXiv:2310.08044},
  year={2023}
}

@inproceedings{tosi2024diffusion,
  title={Diffusion models for monocular depth estimation: Overcoming challenging conditions},
  author={Tosi, Fabio and Ramirez, Pierluigi Zama and Poggi, Matteo},
  booktitle={European Conference on Computer Vision},
  pages={236--257},
  year={2024}
}

@inproceedings{ji2023ddp,
  title={Ddp: Diffusion model for dense visual prediction},
  author={Ji, Yuanfeng and Chen, Zhe and Xie, Enze and Hong, Lanqing and Liu, Xihui and Liu, Zhaoqiang and Lu, Tong and Li, Zhenguo and Luo, Ping},
  booktitle={Proceedings of the IEEE/CVF International Conference on Computer Vision},
  pages={21741--21752},
  year={2023}
}

@inproceedings{duan2024diffusiondepth,
  title={Diffusiondepth: Diffusion denoising approach for monocular depth estimation},
  author={Duan, Yiquan and Guo, Xianda and Zhu, Zheng},
  booktitle={European Conference on Computer Vision},
  pages={432--449},
  year={2024}
}

@article{saxena2023surprising,
  title={The surprising effectiveness of diffusion models for optical flow and monocular depth estimation},
  author={Saxena, Saurabh and Herrmann, Charles and Hur, Junhwa and Kar, Abhishek and Norouzi, Mohammad and Sun, Deqing and Fleet, David J},
  journal={Advances in Neural Information Processing Systems},
  volume={36},
  pages={39443--39469},
  year={2023}
}

@article{hu2024metric3d,
  title={Metric3d v2: A versatile monocular geometric foundation model for zero-shot metric depth and surface normal estimation},
  author={Hu, Mu and Yin, Wei and Zhang, Chi and Cai, Zhipeng and Long, Xiaoxiao and Chen, Hao and Wang, Kaixuan and Yu, Gang and Shen, Chunhua and Shen, Shaojie},
  journal={IEEE Transactions on Pattern Analysis and Machine Intelligence},
  volume={46},
  number={12},
  pages={10579--10596},
  year={2024},
  publisher={IEEE}
}

@inproceedings{yin2023metric3d,
  title={Metric3d: Towards zero-shot metric 3d prediction from a single image},
  author={Yin, Wei and Zhang, Chi and Chen, Hao and Cai, Zhipeng and Yu, Gang and Wang, Kaixuan and Chen, Xiaozhi and Shen, Chunhua},
  booktitle={Proceedings of the IEEE/CVF International Conference on Computer Vision},
  pages={9043--9053},
  year={2023}
}

@inproceedings{choi2022perception,
  title={Perception prioritized training of diffusion models},
  author={Choi, Jooyoung and Lee, Jungbeom and Shin, Chaehun and Kim, Sungwon and Kim, Hyunwoo and Yoon, Sungroh},
  booktitle={Proceedings of the IEEE/CVF Conference on Computer Vision and Pattern Recognition},
  pages={11472--11481},
  year={2022}
}

@inproceedings{lee2025beta,
  title={Beta Sampling is All You Need: Efficient Image Generation Strategy for Diffusion Models Using Stepwise Spectral Analysis},
  author={Lee, Haeil and Lee, Hansang and Gye, Seoyeon and Kim, Junmo},
  booktitle={Proceedings of the IEEE/CVF Winter Conference on Applications of Computer Vision},
  pages={4215--4224},
  year={2025},
  organization={IEEE}
}

@inproceedings{yang2023diffusion,
  title={Diffusion probabilistic model made slim},
  author={Yang, Xingyi and Zhou, Daquan and Feng, Jiashi and Wang, Xinchao},
  booktitle={Proceedings of the IEEE/CVF Conference on Computer Vision and Pattern Recognition},
  pages={22552--22562},
  year={2023}
}

@inproceedings{si2024freeu,
  title={Freeu: Free lunch in diffusion u-net},
  author={Si, Chenyang and Huang, Ziqi and Jiang, Yuming and Liu, Ziwei},
  booktitle={Proceedings of the IEEE/CVF Conference on Computer Vision and Pattern Recognition},
  pages={4733--4743},
  year={2024}
}

@inproceedings{leng2025repa,
  title={Repa-e: Unlocking vae for end-to-end tuning of latent diffusion transformers},
  author={Leng, Xingjian and Singh, Jaskirat and Hou, Yunzhong and Xing, Zhenchang and Xie, Saining and Zheng, Liang},
  booktitle={Proceedings of the IEEE/CVF International Conference on Computer Vision},
  pages={18262--18272},
  year={2025}
}

\begin{IEEEbiography}[{\includegraphics[width=1in,height=1.25in,clip,keepaspectratio]{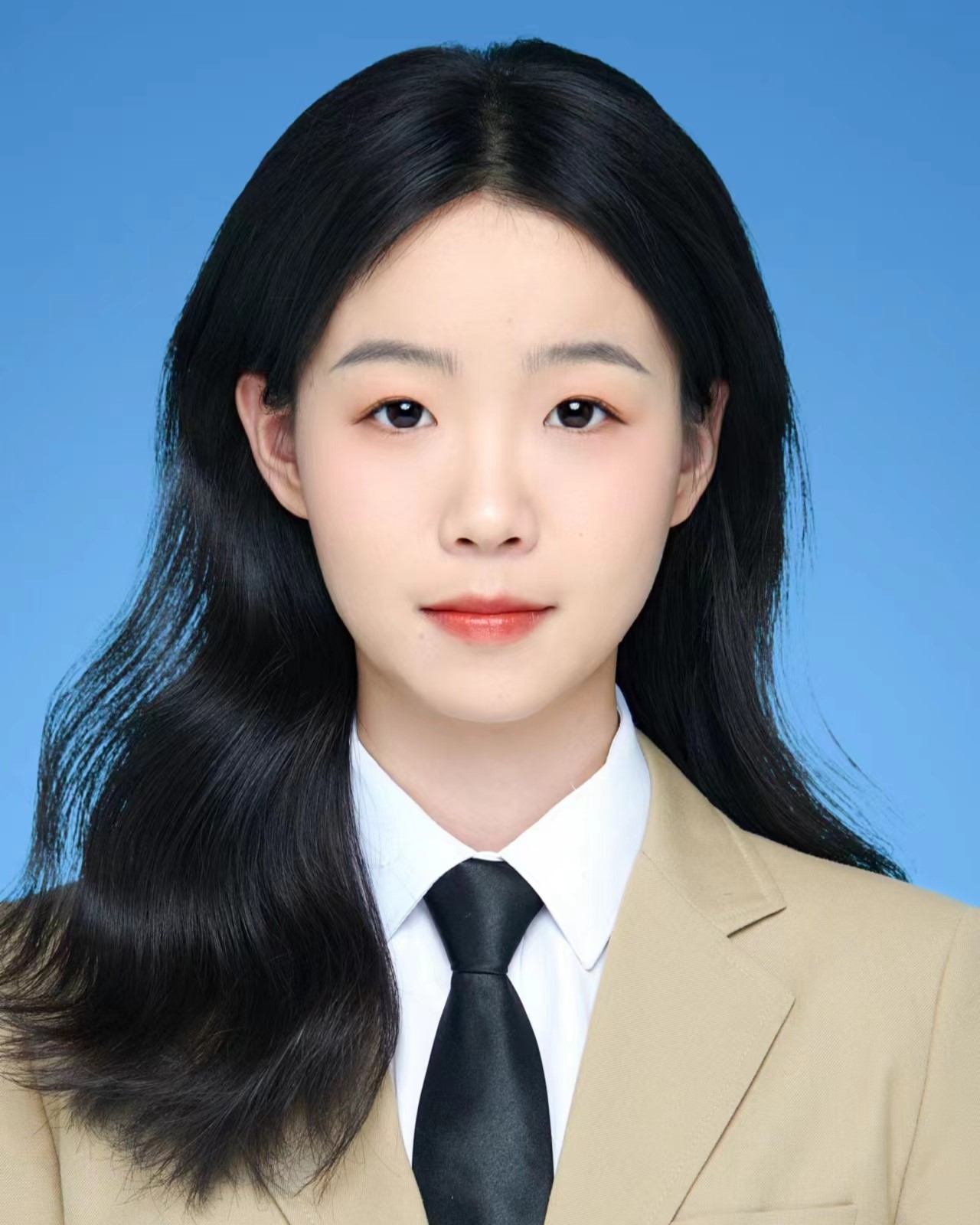}}]{Ziyang Song}
received a bachelor's degree in Control Science and Engineering from University of Science and Technology of China in 2022. She is now pursuing a master's degree in Control Science and Engineering at University of Science and Technology of China. Her research interests include computer vision, especially depth estimation and multi-view stereo.
\end{IEEEbiography}

\begin{IEEEbiography}[{\includegraphics[width=1in,height=1.25in,clip,keepaspectratio]{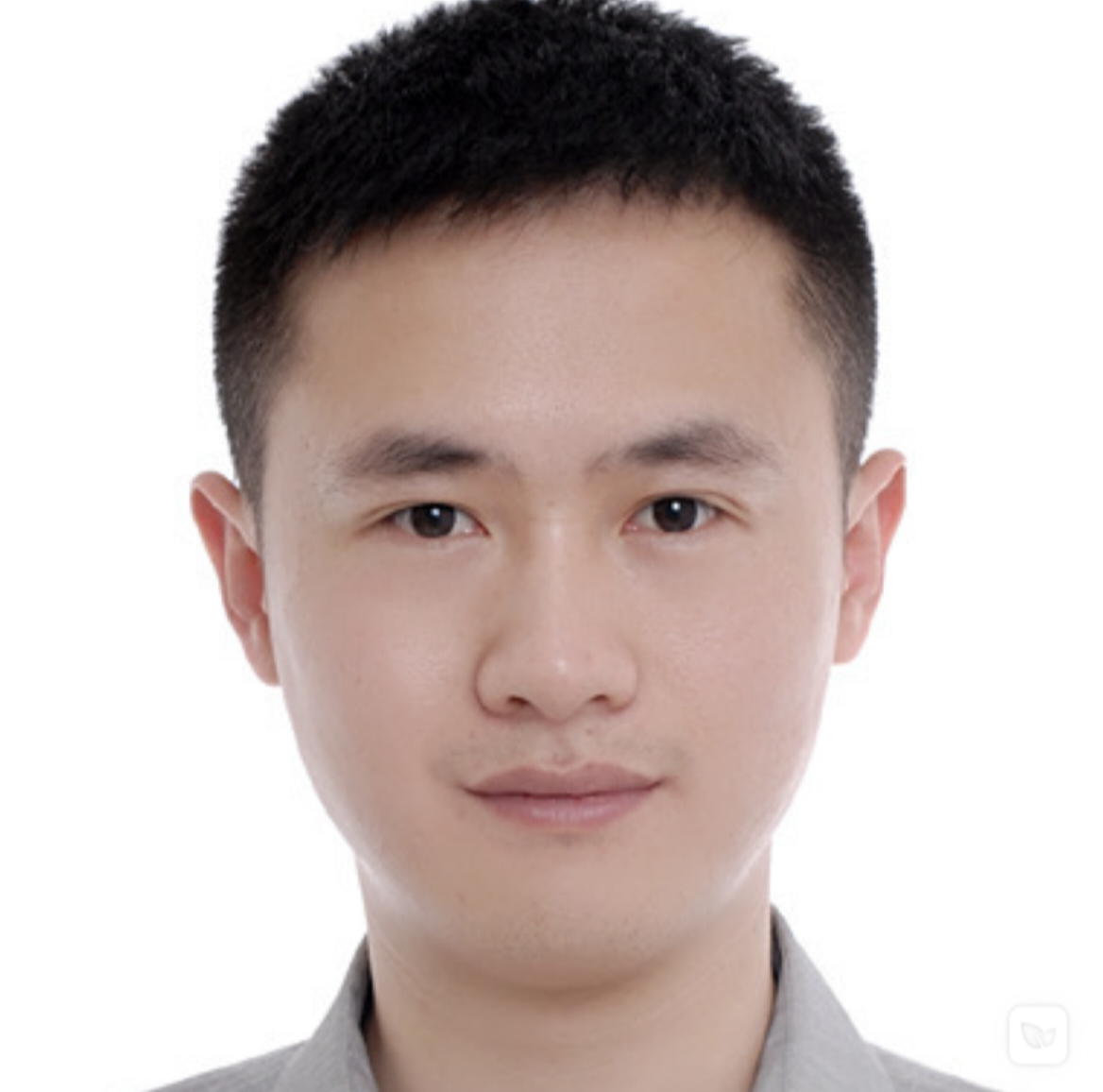}}]{Zerong Wang}
is currently a researcher \& engineer in the Quality Enhancement Center of vivo. Before that,  He received a bachelor's degree in Computer Science from Northwestern Polytechnical University, Xi'an, China, in 2015 and a master's degree in Computer Application Technology from Sichuan University, Chengdu, China, in 2018. His research interests include computer vision, artificial intelligence and computational photography.
\end{IEEEbiography}

\begin{IEEEbiography}[{\includegraphics[width=1in,height=1.25in,clip,keepaspectratio]{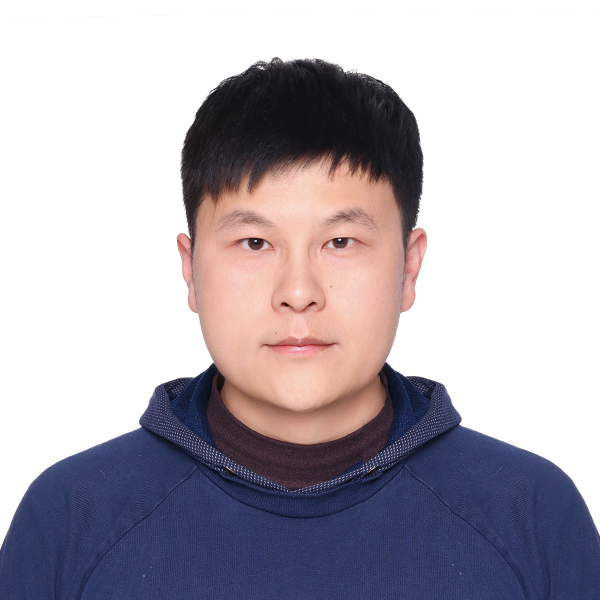}}]{Bo Li}
received the BSc and PhD degrees from the Department of Computer Science, Nanjing University, China, in 2014 and 2019, respectively. From 2020 to 2023, he served as Senior Researcher of Youtu Lab in Tencent, China. He is currently a senior expert at vivo image algorithm research department, China. His research interests include computer vision, pattern recognition and artificial intelligence.
\end{IEEEbiography}

\begin{IEEEbiography}[{\includegraphics[width=1in,height=1.25in,clip,keepaspectratio]{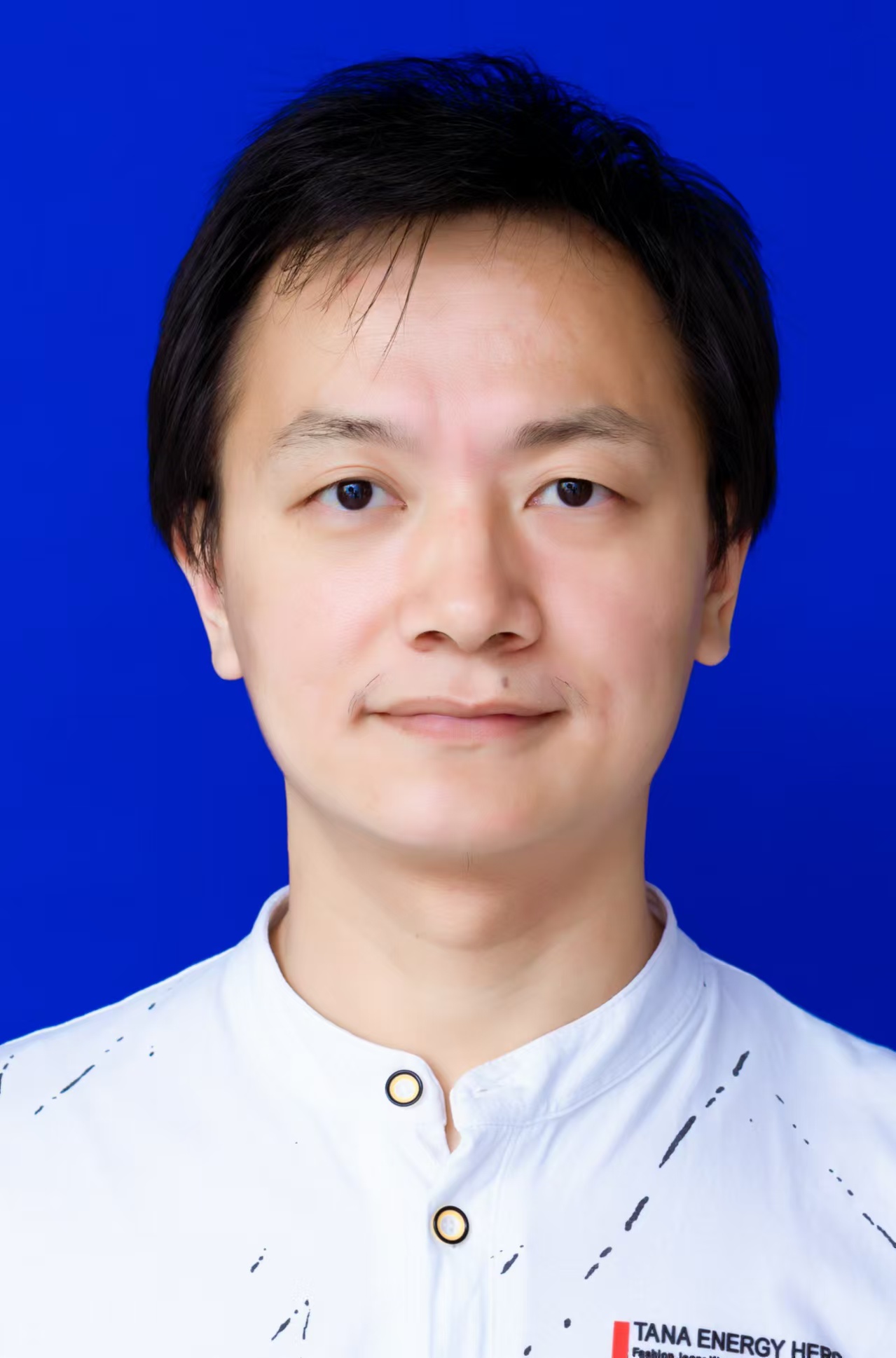}}]{Hao Zhang}
received the BSc and Master degrees from the Beihang University (BUAA), China, in 2012 and 2015, respectively. At the same time, he also obtained the French General Engineer Diploma from Ecole Centrale de Pekin and Ecole Centrale de Lyon.
After that, he served as Senior Researcher of Alipay and Tencent, China. He is currently a Senior Researcher at vivo Mobile Communication Co., Ltd., China. His research interests include computer vision, artificial intelligence and computational photography.
\end{IEEEbiography}

\begin{IEEEbiography}[{\includegraphics[width=1in,height=1.25in,clip,keepaspectratio]{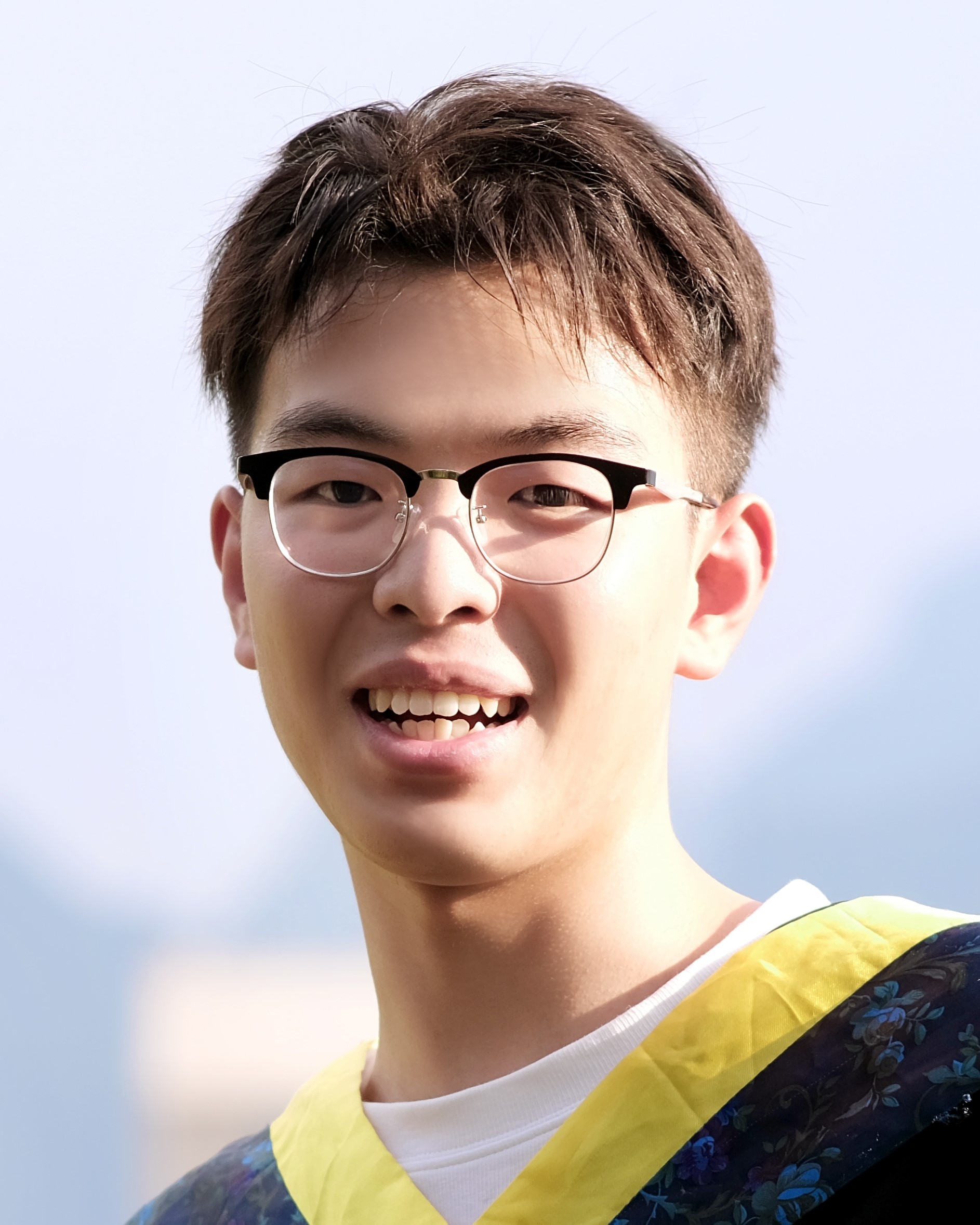}}]{Ruijie Zhu}
received a bachelor's degree in Computer Science from Northwestern Polytechnical University, Xi'an, China, in 2022. 
He is currently pursuing a master's degree in Electronic Engineering
and Information Science at the University of Science and Technology of China. 
His research interests include computer vision and machine learning, especially depth estimation, 3D reconstruction and neural rendering.
\end{IEEEbiography}

\begin{IEEEbiography}[{\includegraphics[width=1in,height=1.25in,clip,keepaspectratio]{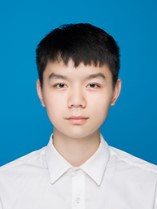}}]{Li Liu}
received the bachelor's degree in Computer Science from Hefei University of Technology, China, in 2022. 
He is currently pursuing a master's degree in Electronic and Information Engineering at the University of Science and Technology of China. 
His research interests include computer vision and machine learning, especially depth estimation and stereo matching.
\end{IEEEbiography}

\begin{IEEEbiography}
[{\includegraphics[width=1in,height=1.25in,clip,keepaspectratio]{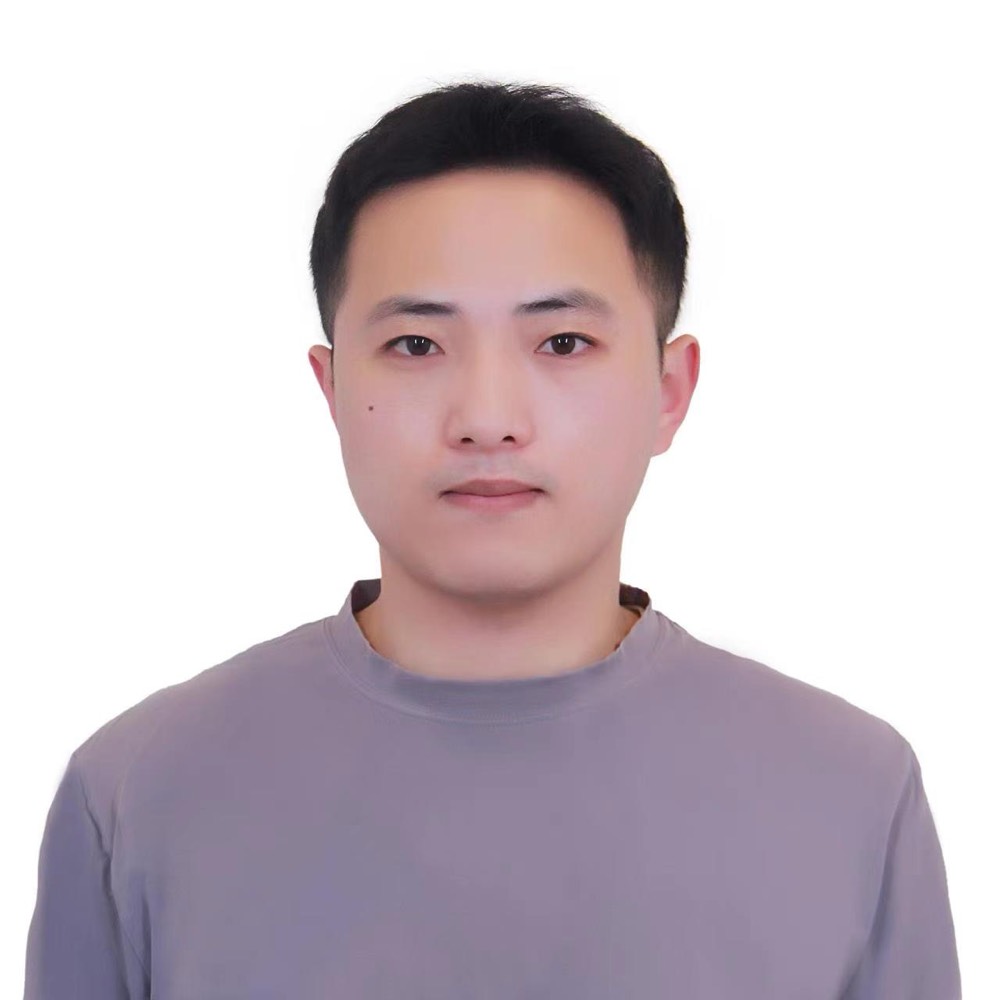}}]{Peng-Tao Jiang}
is currently a lead researcher \& engineer in the Quality Enhancement Center of vivo. Before that, He was a post-doc researcher at Zhejiang University, working with Prof. Chunhua Shen. He received his PhD at Nankai University, advised by Prof. Ming-Ming Cheng. His current research interests include diffusion, image restoration, multi-task learning, and segmentation.
\end{IEEEbiography}

\begin{IEEEbiography}
[{\includegraphics[width=1in,height=1.25in,clip,keepaspectratio]{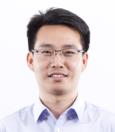}}]{Tianzhu Zhang}
received a bachelor's degree in communications and
information technology from the Beijing Institute
of Technology, Beijing, China, in 2006 and a Ph.D.
in pattern recognition and intelligent systems from
the Institute of Automation, Chinese Academy of
Sciences, Beijing, China, in 2011. He is currently
a Professor at the Department of Automation,
School of Information Science and Technology,
University of Science and Technology of China.
His current research interests include computer
vision and multimedia. He served/serves as the Area Chair for CVPR 2020,
ECCV 2020, ICCV 2019, ACM MM 2019, WACV 2018, ICPR 2018,
and MVA 2017, the Associate Editor for IEEE T-CSVT and Neurocomputing.
\end{IEEEbiography}

\vfill

\end{document}